%% file: main-acl.tex
\documentclass[11pt]{article}

\usepackage[final]{acl}

\usepackage{newtxtext}
\usepackage{newtxmath}
\usepackage{latexsym}
\usepackage{fix-cm}
\usepackage[T1]{fontenc}

\usepackage[utf8]{inputenc}

\usepackage{microtype}

\usepackage{inconsolata}

\usepackage{graphicx}

\usepackage{multicol}
\usepackage{multirow}

\usepackage[most]{tcolorbox} 
\usepackage{listings}        
\usepackage{xcolor}          
\usepackage{enumitem} 
\usepackage{comment}  
\usepackage{graphicx} 
\usepackage{threeparttable}  
\usepackage{booktabs}  
\usepackage{makecell}  
\usepackage{pifont} 
\usepackage[frozencache=true,cachedir=_minted]{minted} 
\usepackage{capt-of} 
\usepackage{adjustbox}
\usepackage[ruled,vlined]{algorithm2e} 
\usepackage{subcaption}

\definecolor{myviolet}{RGB}{133, 96, 136}

\newtcolorbox{violetlisting}[1][]{%
	listing only,
	listing options={
		basicstyle=\ttfamily\small,
		breaklines=true,      
		breakatwhitespace=false,
		columns=fullflexible,
		keepspaces=true,
		language=Python       
	},
	colback=myviolet!5,
	colframe=myviolet,
	title=Code Example,
	fonttitle=\bfseries,
	sharp corners,
	enhanced,
	breakable,
	#1
}

\definecolor{systemcolor}{HTML}{888888}    
\definecolor{usercolor}{HTML}{1F77B4}      
\definecolor{assistantcolor}{HTML}{2CA02C} 
\definecolor{toolcolor}{HTML}{D62728}      
\definecolor{callcolor}{HTML}{9467BD}      

\newcommand{\systemline}[1]{\textcolor{systemcolor}{\textbf{System:} #1}}
\newcommand{\userline}[1]{\textcolor{usercolor}{\textbf{User:} #1}}
\newcommand{\assistantline}[1]{\textcolor{assistantcolor}{\textbf{Assistant:} #1}}
\newcommand{\toolcall}[1]{%
	\textcolor{callcolor}{\textbf{Assistant:}\\ \begin{minipage}[t]{0.9\linewidth}\ttfamily\raggedright #1\end{minipage}}%
}

\newcommand{\toolresp}[1]{%
	\textcolor{toolcolor}{\textbf{Tool:}\\ \begin{minipage}[t]{0.9\linewidth}\ttfamily\raggedright #1\end{minipage}}%
}

\setminted{
	breaklines=true,
	breakanywhere=true, 
	fontsize=\small,
	autogobble=true,    
	tabsize=2,
	frame=single,
	framesep=2mm,
	bgcolor=gray!5
}

\newcommand{\cmark}{\textcolor{green}{\ding{51}}} 
\newcommand{\xmark}{\textcolor{red}{\ding{55}}}   

\title{ToolWeave: Structured Synthesis of Complex Multi-Turn Tool-Calling Dialogues}

\author{Dinesh Khandelwal$^{1}$, Gnana Prakash Punnavajhala$^{2}$\thanks{Work done during Gnana's internship at IBM Research India.}, GPS Bhargav$^{1}$, Gaurav Pandey$^{1}$,\\ {\bf Sachin Joshi$^{1}$, Hima Karanam$^{1}$, Dinesh Raghu$^{1}$}\\
	$^{1}$IBM Research, $^{2}$IIIT Hyderabad\\
	$^{1}$\texttt{\{dikhand1, gpshri27, gpandey1, jsachind, hkaranam, diraghu1\}@in.ibm.com},\\
	$^{2}$\texttt{punnavajhala.prakash@research.iiit.ac.in}
}

\begin{document}
\maketitle
\input{sections/abstract}
\input{sections/intro}
\input{sections/related}
\input{sections/architecture}
\input{sections/data-analysis}
\input{sections/expt_new}
\input{sections/conclusion}

\bibliography{custom}

\appendix
\clearpage
\input{sections/appendix_two_column}

\end{document}

%% file: sections/abstract.tex
\begin{abstract}
Multi-turn tool calling is essential for LLMs to function as autonomous agents, yet synthesizing the training data required for these capabilities remains a fundamental challenge. Existing synthetic data generation pipelines often produce unrealistic dialogues for two reasons: they chain tools that are only superficially compatible rather than aligned with meaningful user tasks, and they generate dialogues in one shot, which often introduces arguments that were neither provided by the user nor produced by prior tool calls. These issues also lead to a severe underrepresentation of multi-step tool interactions.
We introduce \emph{ToolWeave}, a structured framework for synthesizing realistic multi-turn tool-calling dialogues. ToolWeave support realistic multi-step workflows (or tool sequences) by constructing tools with built-in dependencies and filters the workflows based on alignment with user goals. It reduces parameter hallucination by using a fine-grained planning stage that explicitly tracks parameter provenance. As a result, ToolWeave-generated synthetic dialogues contain more multi-step tool interactions (45\%) and fewer hallucinations in parameters and tool names. Consequently, LLMs fine-tuned on \emph{ToolWeave} consistently outperform those fine-tuned on prior datasets across three public benchmarks. Notably, \texttt{Llama-3.1-70B} fine-tuned on ToolWeave achieves 39.75\% on BFCL-V3 multi-turn, compared to 23.50\% when fine-tuned on SOTA \emph{ToolFlow} data.
\end{abstract}

%% file: sections/intro.tex
\section{Introduction}\label{sec:intro}

The evolution of large language models (LLMs) into autonomous agents relies on \emph{multi-turn tool-calling}: the ability to chain tools, elicit user information, and maintain state across complex workflows \citep{yao2023react, toolformer-2023-neurips}. Fine-tuning on synthetic data has become a standard approach for improving these capabilities \citep{qin2024toolllm}. However, generating realistic multi-turn tool-calling dialogues is not simply generating fluent conversations with tool calls. Existing synthesis pipelines, such as ToolFlow~\cite{shim2025tooldial}, Magnet~\cite{yin-etal-2025-magnet}, and ToolDial~\cite{shim2025tooldial}, suffer from two major limitations: semantically incoherent workflows and parameter hallucinations.

\noindent \textit{Semantically incoherent workflows:} Given a set of tools, existing approaches construct a tool graph and sample sequences of connected tools (i.e., workflows) for dialogue synthesis. However, the coherence of these workflows is fundamentally limited by the quality of the underlying tool graph. In most pipelines, edges in the tool graph are inferred by matching parameter names and descriptions across tools. For instance, a pipeline may connect \texttt{create\_support\_ticket(issue)}~$\rightarrow$~\texttt{ticket\_id} with \texttt{cancel\_movie\_ticket(ticket\_id)} solely because both expose a parameter named \texttt{ticket\_id}. While this enforces schema-level compatibility, it does not guarantee semantic coherence.

This issue has an important downstream consequence. Because workflows are constructed based on superficial similarity rather than true data dependencies, they rarely require the output of one tool to meaningfully drive another.
As a result, the synthesized dialogues seldom contain genuine \emph{multi-step} tool interactions within a turn. This is a crucial limitation, as real world tool use tasks require composing multiple dependent operations, where intermediate results must be propagated across tools to achieve a single user objective. In our analysis (Table~\ref{tab:dialogue_stats}), fewer than 7\% of dialogues from prior pipelines contain multi-step dialogues.

\noindent \textit{Parameter hallucinations:} Given a candidate workflow, existing approaches typically prompt an LLM to generate the entire dialogue or  plan in a single pass. This leaves the execution structure implicit. The model must infer how tool calls align with user intent, where each argument originates from, and how intermediate results are propagated. In practice, this often leads to hallucinated arguments, where parameters are neither provided by the user nor produced by preceding tool calls. As shown in Table~\ref{tab:hallucination}, more than 50\% of dialogues generated by prior approaches contain such parameter hallucinations, and models trained on this data risk inheriting the same behaviour.

These limitations stem from a common root cause: the absence of structured, verifiable intermediate representations that explicitly capture workflow dependencies and parameter provenance.

\noindent \textbf{Our Approach:} To address these challenges, we introduce \emph{ToolWeave}, a modular framework that decomposes synthesis into structured, verifiable stages. ToolWeave is designed to generate dialogues with realistic multi-step tool interactions while minimizing hallucinations.

First, instead of starting with a fixed set of tools and inferring edges in the tool graph post hoc via semantic similarity, we propose an iterative curriculum that synthesizes tools with built-in information flow between them. This yields tool graphs with meaningful data dependencies and coherent workflows by construction. We further improve workflow quality through an explicit verification step. We sample workflows, synthesize user goals for each workflow, score the compatibility between workflows and goals, and retain only high-quality pairs, thereby filtering out incoherent compositions. This increases the prevalence of multi-step tool interactions from $\sim$7\% in prior approaches to  $\sim$45\%.

Second, we introduce a fine-grained planning stage that explicitly models \emph{parameter provenance}.  
For each dialogue, the plan specifies parameters at each step and their sources (user, context, or prior tools).
This explicit structure eliminates ambiguity in execution and significantly reduces hallucinated arguments, lowering hallucination rates from $\sim$53\% in prior approaches to nearly 20\%.

ToolWeave consistently outperforms existing approaches across three publicly available tool-calling benchmarks. 

%% file: sections/related.tex
\section{Related Work}\label{sec:related}

\paragraph{Fine-Tuning Datasets and Models for Tool Calling:}  
Toolformer~\citep{toolformer-2023-neurips}, replaced text segments with tool calls to train LLMs to produce tool calls. Several instruction-tuning datasets have been created to improve tool-calling capabilities in LLMs. Most of these datasets introduce single-turn tool-calling capabilities in models~\citep{apigen-2024-neurips, nestful, apiblend-2024-acl, toolAlpaca-2025-www,qin2024toolllm} using APIs extracted from sources such as  RapidAPI~\citep{rapidapi} or APIBench~\citep{APIbench-gorilla-2024-nuerips}. Reinforcement learning and other tuning strategies to improve tool calling capabilities are presented in~\citet{ExploringSuperiorFunctionCalls, chen-etal-2025-enhancing}. Tool retrieval becomes an important aspect with increasing tools and is addressed in the works~\citet{qin-etal-2025-meta,chen-etal-2024-invoke}.
Recent work explores looking at creating multi-turn tool calling data using synthetic data generation pipelines and models~\citep{wang2025toolflow, shim2025tooldial, apigen-MT2025, toolAce2025iclr, yin-etal-2025-magnet,button-2025-iclr}. 
The APIs and frontier models used in these pipelines make them restrictive and not licensable for commercial deployments. Random sampling of APIs followed by high-level plan generation used in these approaches often results in incoherent and poor-quality dialogues. To address these shortcomings, ToolWeave proposes an end-to-end synthetic data generation pipeline that provides fine-grained control at each stage that works equally well with frontier and small open-weight models.

\paragraph{Tool Calling Evaluation Benchmarks:}
BFCL~\citep{patil2025bfcl} is one of the comprehensive benchmarks aimed at evaluating various aspects of function calling including multi-turn. Several other benchmarks evaluate single-turn capabilities of the models like~\citep{sealtools, patil2025bfcl, when2call-2025-naacl, toolAlpaca-2025-www,xu2023toolbench,zhuang2023toolqa, metatool-2024-iclr,nestful}. Benchmarks like CONFETTI~\citep{alkhouli-etal-2025-confetti}, APIBank~\citep{li-etal-2023-apibank}, ToolHop~\cite{ye2025toolhop} are built for evaluating multi-turn capabilities of the models. $\tau$-bench~\citep{tauBench-2025-iclr} and Agentboard~\citep{agentboard-2024-neurips} assess agentic capabilities, including multi-turn tool use and policy adherence.

Appendix~\ref{sec:sdg_pipeline_comparsion} compares ToolWeave with other SDG pipelines across key features, including code availability, API synthesis, licensing, and fine-grained control.

%% file: sections/architecture.tex
\section{ToolWeave Architecture}
\label{sec:architecture}

\begin{figure*}[ht]
	\centering
	\includegraphics[width=\linewidth]{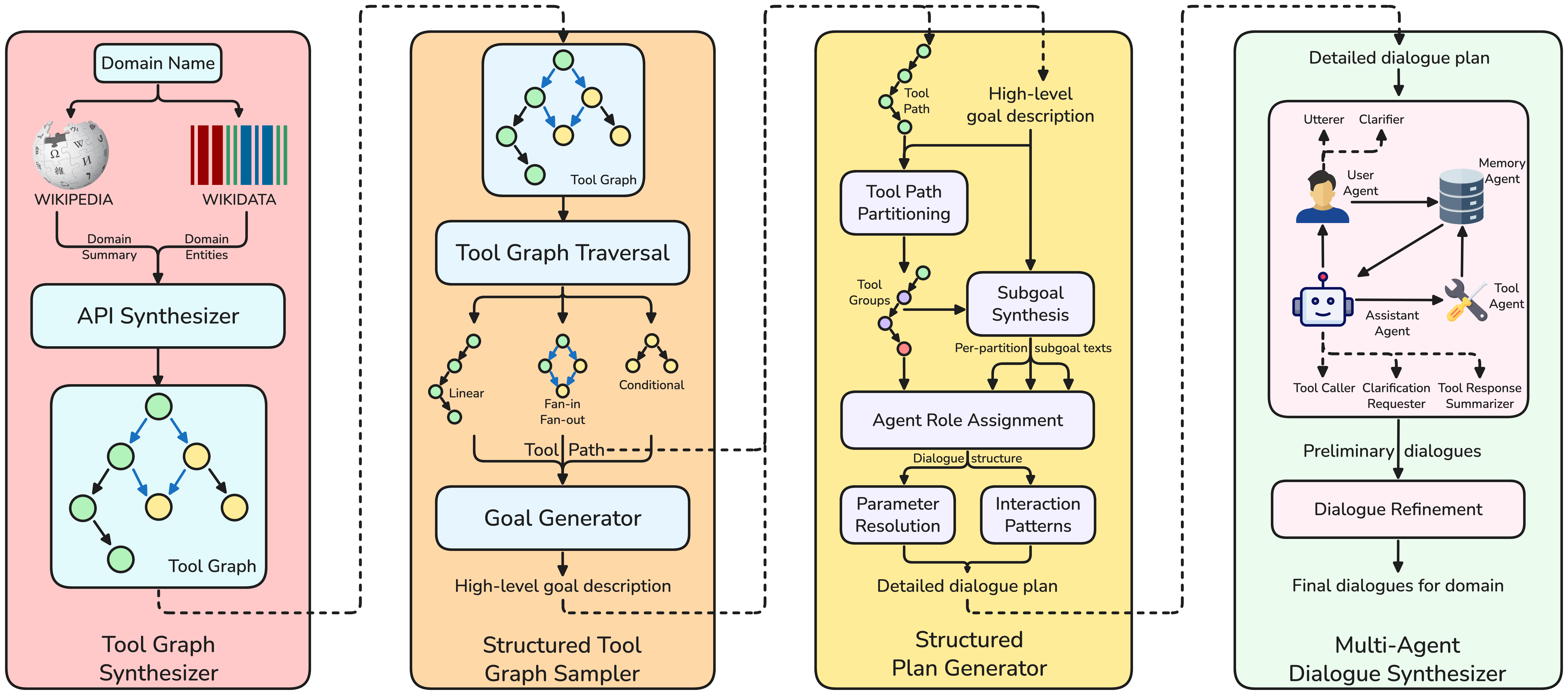}
	\caption{The modular architecture of ToolWeave. Starting with a domain name, the \emph{Tool Graph Synthesizer} creates synthetic APIs and organizes them into an interconnected graph; the \emph{Structured Tool Graph Sampler} extracts coherent subgraphs as goals; the \emph{Structured Plan Generator} expands these goals into detailed dialogue plans; and the \emph{Multi-Agent Dialogue Synthesizer} instantiates the plans into natural multi-turn dialogues.}
	\label{fig:toolweave_architecture}
\end{figure*}

ToolWeave is a modular pipeline for synthesizing realistic multi-turn, multi-step tool-calling dialogues from automatically generated tools. It decomposes dialogue generation into four stages that progressively transform structured intermediate representations into natural conversations. As shown in Figure~\ref{fig:toolweave_architecture}, ToolWeave first constructs a synthetic tool graph, then samples goal-directed workflows, converts them into fine-grained executable plans, realizes these plans as multi-agent dialogues with grounded tool calls, and finally applies post-processing to introduce linguistic variation and controlled noise. By externalizing complex reasoning into these structured stages, ToolWeave enables reliable generation of high-quality tool-calling trajectories. A full ToolWeave example (goal, plan, tools, and dialogue) is shown in Appendix~\ref{app:example}.

\subsection{Tool Graph Synthesizer}
\label{sec:tool_graph_synthesizer}

\begin{figure}[ht]
	\centering
	\includegraphics[width=\linewidth]{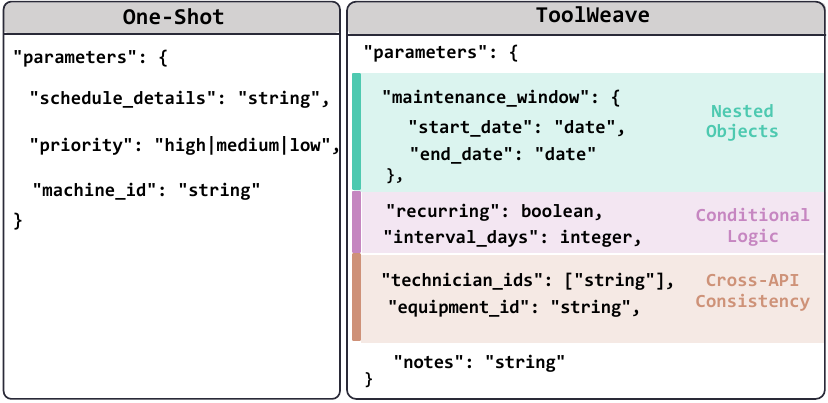}
	\caption{Input schema comparison for \texttt{generate\_maintenance\_schedule()}. One-shot generation yields a flat parameter list, while ToolWeave produces a richer schema with nested objects.}
	\label{fig:apis}
\end{figure}

Prior tool graph construction typically relies on either real-world APIs from marketplaces like RapidAPI  \citep{rapidapi} or one-shot synthetic tool generation via LLMs. The RapidAPI-based approaches provide realistic tools but have two limitations. First, existing frameworks typically infer connections between tools using semantic similarity over parameter names and descriptions \citep{shim2025tooldial, wang2025toolflow}. Such heuristics often create incorrect edges; e.g., unrelated parameters like \texttt{get\_hotel\_id(.)~$\to$~id} and \texttt{reserve\_restaurant(id)} may be falsely linked as compatible, leading to invalid tool compositions. Second, many tools lack clear redistribution rights, raising potential licensing concerns.

Synthetic generation approaches \citep{toolAce2025iclr, button-2025-iclr} avoid licensing issues but generate all tools in a single step. As a result, the generated tools tend to cover only superficial functionality, rarely capturing deeper schema relationships, real-world business logic, or workflows across tools. 

In contrast, ToolWeave's multi-stage curriculum produces highly structured and realistic definitions. Figure~\ref{fig:apis} illustrates this contrast by comparing the input parameter schemas for a \texttt{generate\_maintenance\_schedule()} tool. While the one-shot baseline generates a flat, simplistic list of parameters, ToolWeave generates complex structures such as nested objects, conditional parameter dependencies, and explicit identifiers that link to other tools. This is made possible by a curriculum-driven, multi-stage tool generation process that ToolWeave follows, described next.

\begin{enumerate}[leftmargin=*,noitemsep,topsep=0pt]
\item \textit{Seed generation:} Create a minimal set of core tools covering fundamental operations.
\item \textit{Entity expansion:} Expand coverage using structured knowledge from Wikipedia and Wikidata. E.g., ToolWeave captures the business rule that \texttt{interval\_days} is mandatory when \texttt{recurring} is true. This is inferred from Wikipedia's maintenance page (see Figure~\ref{fig:apis}).
\item \textit{Schema enrichment:} Add depth via nested objects, enums, defaults, and optional parameters.
\item \textit{Connection discovery:} Introduce explicit linking parameters that enable workflows across tools (e.g., \texttt{technician\_id} and \texttt{equipment\_id}), rather than inferring edges post-hoc via semantic similarity.
\end{enumerate}

By decomposing tool synthesis into staged steps, ToolWeave constructs a domain-specific tool graph $\mathcal{G}$, where nodes represent tools and directed edges indicate that outputs of one tool may serve as inputs to another. This graph captures explicit data dependencies between tools and serves as the foundation for workflow sampling in the next stage.  More details on the Tool Graph Synthesizer are provided in Appendix~\ref{app:synthethic_apis_details}.

\subsection{Structured Tool Graph Sampler}
\label{sec:goal_generation}

Given the synthesized tool graph, the next step is to sample coherent tool workflows aligned with meaningful user goals. Prior work~\citep{wang2025toolflow} generates tool sequences using random walks over tool graphs. While this ensures syntactic compatibility between tools, it often produces sequences that do not correspond to meaningful user tasks.  ToolWeave instead enforces \emph{tool–goal alignment} by sampling workflows that follow common structural motifs: \emph{linear}, \emph{fan-in/fan-out}, and \emph{conditional}. These motifs mimic real-world tool interactions and enable the generation of goal-directed tool sequences. By constraining sampling to such motifs, ToolWeave transforms tool-sequence generation from an unconstrained graph traversal into a controlled workflow synthesis process.

\paragraph{Workflow motifs:}
ToolWeave samples workflows using common structural motifs: \emph{linear}, \emph{fan-in/fan-out}, and \emph{conditional}. Linear chains form the backbone of the sampler, reflecting typical interactions where each step builds on previous results (e.g., search tickets $\to$ retrieve details $\to$ update status $\to$ notify user). These chains are obtained through a random walk over the tool graph.  Higher-order motifs capture more complex patterns. \emph{Fan-in/fan-out} workflows aggregate multiple pieces of information before a final action (e.g., retrieving ticket details and checking user history before updating the ticket). \emph{Conditional} workflows model branches where the next step depends on an API result (e.g., retrieve ticket details if creation succeeds, otherwise list existing tickets).

\paragraph{Validation and Selection:}
For each sampled workflow, \emph{ToolWeave} synthesizes a natural-language goal and scores the pair $(\text{tools in workflow}, \text{goal})$ using an LLM judge. After discarding pairs below a minimum quality threshold, we apply Maximum Marginal Relevance (MMR)~\citep{carbonell1999mmr} to the linear workflows to prune similar goals and maximize task diversity. For \emph{fan-in/fan-out} and \emph{conditional} motifs, all valid pairs are retained to preserve complex interaction patterns. Detailed scoring rubrics and algorithms are provided in Appendix~\ref{app:goal_gen_details}.

\subsection{Structured Plan Generator}
\label{sec:dialogue_planner}

Given a goal and its tool sequence, the next step is to generate a plan that guides dialogue synthesis. Broadly, plan generation strategies fall into two categories: high-level planning and turn-level planning. High-level plans~\cite{wang2025toolflow} specify only abstract steps, leaving most work to the dialogue synthesizer. While flexible, this places a heavy burden on synthesizer and often leads to errors such as repeated interaction loops (e.g., repeated goodbyes), incorrect tool ordering (e.g., shipping an order before it is created), or hallucinated outputs produced without corresponding tool calls. In contrast, turn-level plans explicitly specify the structure of each dialogue turn. However, existing turn-level approaches~\cite{shim2025tooldial} rely on predefined action-sequence templates that do not explicitly model parameter dependencies across tools, and rarely generate dialogues containing multi-step tool interactions within the same turn, producing such patterns in only ~7\% of dialogues.

To address these limitations, ToolWeave shifts much of the reasoning burden from the dialogue synthesizer to an explicit planning stage, so that the synthesizer mainly realizes the dialogue from the plan rather than performing complex reasoning.  Unlike prior methods that infer dialogue structure during generation, ToolWeave computes tool grouping, parameter dataflow, and clarification requirements during planning. As a result, the planner enables richer interaction patterns, increasing the proportion of dialogues containing multi-step tool interactions within a single turn to nearly 50\%. The planner operates in three stages.

\paragraph{Tool Path Partitioning:}
Given a goal and its corresponding tool sequence (e.g., [A, B, C, D]), the planner partitions the sequence into contiguous groups (e.g., [[A], [B, C], [D]]). Each group represents a coherent sub-goal realized as a dialogue segment. Linear paths are partitioned by an LLM conditioned on the goal to identify natural subgoal boundaries. Higher-order motifs use deterministic rules to preserve dependencies between tools.

\paragraph{Subgoal Synthesis and Segmentation:}

Next, each tool group is assigned a concise natural-language subgoal, and the corresponding dialogue segment is constructed. Each segment is realized as a sequence of interaction acts including \texttt{user\_request}, \texttt{assistant\_clarification}, \texttt{user\_clarification\_response}, \texttt{tool\_invocation}, and \texttt{assistant\_response}. A segment begins with a user request expressing the subgoal. If required parameters are missing, the assistant issues clarification requests, followed by user responses that provide the requested inputs, before executing the corresponding tool calls and summarizing the results of the tool executions.

\paragraph{Parameter Planning:}
Within each segment, the planner resolves tool parameters by tracing the Tool Graph. Parameters are categorized as (i) user-provided (from user utterances) and (ii) derivable (populated from prior tool outputs or dialogue history). This explicitly plans dataflow instead of relying on the dialogue synthesizer to infer dependencies during generation.

Overall, the Structured Plan Generator introduces three key design elements. Tool grouping partitions the tool path into tool groups, enabling multi-step tool interactions within a dialogue segment. Fine-grained interaction acts improve conversational coherence. Explicit parameter binding ensures consistent data flow across turns. Together, these choices transform dialogue generation from a loosely guided language generation problem into a structured execution process, enabling reliable synthesis of complex multi-turn tool-calling dialogues. Planner algorithms are described in Appendix~\ref{app:dialogue_planner}.

\subsection{Multi-Agent Dialogue Synthesizer and Post-Processing}\label{sec:dialogue_synthesizer}

Using the structured plan, ToolWeave synthesizes the dialogue using a plan-driven multi-agent framework. Each interaction act is handled by a dedicated agent, along with a tool agent and a memory agent. The memory agent maintains a persistent state containing resolved parameters, tool outputs, and other dialogue facts, ensuring consistency across turns. The tool agent generates schema-consistent responses for tool calls without requiring access to live API endpoints.  The agent responsible for the current interaction act receives the dialogue history, the parameter plan, and the memory state, and produces the next utterance accordingly. The memory is then updated before executing the next step. This stateful execution ensures that dialogues faithfully follow the  plan while maintaining coherent multi-turn interactions. Details are in Appendix~\ref{app:dialogue_synthesis}.

To ensure robustness, we further refine the synthesized dialogues through three post-processing steps: (i) \emph{Linguistic Variation:} rephrasing user utterances to prevent overfitting; (ii) \emph{Failure Injection:} introducing erroneous tool calls and recovery behaviors to simulate real-world API faults; and (iii) \emph{Schema Masking:} replacing function names with generic identifiers (e.g., \texttt{func\_01}) to force reasoning over rich schemas rather than surface labels. This refinement process transforms raw synthetic traces into a natural, error-robust dataset suited for generalizable tool-calling. (Detailed algorithms are in Appendix~\ref{app:robustness_algos}).

%% file: sections/data-analysis.tex
\section{Analysis of Data Quality}

\subsection{API Quality and Coverage}
\label{sec:api_quality}

We evaluate the synthetic APIs generated by ToolWeave along three dimensions: \emph{breadth}, \emph{depth}, and \emph{connectivity}. Breadth measures API diversity via the average number of APIs per domain and parameters per API. Depth captures schema complexity using \emph{Complex API Use (CAU)}—the proportion of APIs containing nested objects or arrays—and the \emph{Required Parameter Ratio (RPR)}, the fraction of parameters marked as required. Connectivity measures how APIs compose into workflows using \emph{Interconnectivity (IC)}, which estimates direct data-flow potential between APIs, and the average longest tool chain. Formal definitions of these metrics are provided in Appendix~\ref{app:api_metrics}.

We generate API sets for 20 domains (Appendix~\ref{app:domain_list}) using three synthesis LLMs: the open-weight \texttt{GPT-OSS-120B}, the commercial \texttt{mistral-medium-2505}, and the frontier \texttt{GPT-5-mini}. Table~\ref{tab:api_stats} reports the per-domain averages. Open-weight models produce broader schemas, while stronger models generate more complex APIs with higher nesting. Despite these stylistic differences, ToolWeave consistently produces coherent and highly connected Tool Graphs.

\setlength{\tabcolsep}{2.5pt}
\begin{table}[t]
	\centering
	\caption{\small Avg. API stats per domain across synthesis seeds.}
	\label{tab:api_stats}
	\scriptsize
	\begin{tabular}{lccc}
		\toprule
		\textbf{Metric} & \textbf{GPT-OSS (O)} & \textbf{Mistral (M)} & \textbf{GPT-5-mini (G)} \\
		\midrule
		Avg. APIs / Domain & 25.25 & 20.85 & 28.35 \\
		Avg. Params / API & 3.60 & 2.92 & 5.01 \\
		Complex API Use (CAU) & 23.6\% & 19.8\% & 36.6\% \\
		Required Param Ratio (RPR) & 65.6\% & 68.4\% & 48.9\% \\
		Interconnectivity (IC) & 1.08 & 1.85 & 1.03 \\
		Longest Chain (avg.) & 5.5 & 5.9 & 6.15 \\
		\bottomrule
	\end{tabular}
\end{table}

\paragraph{Comparison with Real-World APIs:} 
Compared to the 4,474 RapidAPI schemas used in ToolDial~\citep{shim2025tooldial}, our synthetic APIs exhibit higher structural complexity (avg.\ 3.6--5.0 vs.\ 2.4 parameters; CAU up to 36.6\% vs.\ 1.0\%). We argue that this difference is beneficial: higher schema complexity introduces richer parameter structures and constraints, requiring LLMs to perform more precise parameter grounding and composition within each tool call. This intra-tool complexity provides the necessary hooks for multi-step reasoning, as reflected in Section~\ref{subsec:dialogue_structure}, where our dialogues achieve up to 44.87\% multi-step turns versus $\sim$7\% in baseline datasets.  Moreover, complex APIs with expressive parameter spaces provide richer supervision signals for learning tool usage and parameter grounding. As shown in Section~\ref{subsec:benchmark_performance}, these structural properties lead to superior multi-turn tool-calling performance.

\subsection{Dialogue Structure and Complexity}
\label{subsec:dialogue_structure}

For controlled comparison, we analyze $\sim$3.2k ToolWeave dialogues and compare them with matched $\sim$3.2k subsets from the ToolFlow and ToolDial baselines. Table~\ref{tab:dialogue_stats} summarizes the structural properties of these dialogues. We report the number of conversational \emph{turns}, total \emph{tool calls}, and the proportion of \emph{multi-step turns} (turns requiring multiple tool calls). We further distinguish \emph{True multi-step turns}, where later calls directly consume outputs from earlier ones within the same turn. The results highlight a clear difference in generation behavior. ToolFlow produces long dialogues (up to 10 turns and 15 tool calls) but achieves 0.0\% True multi-step turns, indicating uncontrolled sequences of shallow, independent tool calls rather than meaningful reasoning chains. ToolDial~\citep{shim2025tooldial}, despite using GPT-4o with real APIs, generates shallow conversations averaging only $\sim$1.5 tool calls per dialogue and no true multi-step turns. In contrast, ToolWeave produces compact dialogues that are dense with reasoning, achieving up to 40.67\% True multi-step turns. These results suggest that structured planning and tool graph reasoning enable richer multi-step interactions than existing pipelines. Example failure cases from ToolFlow and ToolDial are shown in Appendix~\ref{app:toolflow_example} and Appendix~\ref{app:tooldial_example_dialogues}.

\setlength{\tabcolsep}{2.3pt}
\begin{table}[t]
	\centering
	\caption{\small Statistics of synthetic dialogues grouped by synthesis pipeline.}
	\label{tab:dialogue_stats}
	\scriptsize
	\begin{tabular}{lccc}
		\toprule
		\textbf{Dataset} & \textbf{Min/Max/Avg.} & \textbf{Min/Max/Avg.} & \textbf{\% Total / \% True} \\
		& \textbf{Turns} & \textbf{Tool Calls} & \textbf{Multi-step Turns} \\
		\midrule
		
		ToolWeave & 1/8/2.71 & 1/5/3.22 & 44.87 / 40.67 \\
		
		ToolFlow  & 1/10/6.28 & 0/15/2.20 & 6.77 / 0.0 \\

		ToolDial & 2/6/4.48 & 0/2/1.51 & 4.47 / 0.0 \\
		
		\bottomrule
	\end{tabular}
\end{table}

\subsection{Hallucinations in Synthesized Dialogues}

We randomly sampled 200 dialogues from ToolWeave, ToolDial, and ToolFlow to evaluate dataset quality. Table \ref{tab:hallucination} reports different hallucination types and the percentage of dialogues exhibiting each error. A dialogue may contain multiple error types and thus appear in multiple rows, while the final row reports dialogues with no errors. The results highlight the benefits of detailed parameter planning in ToolWeave. The only hallucinations that persist in ToolWeave are value hallucinations. For example, when a user says \textit{Charge the \$201.40 to my credit card}, the generated tool call may record \textit{payment\_amount: 20140}.

\begin{table}[t]
	\centering
	\scriptsize
	\caption{Hallucination analysis across ToolWeave, ToolDial, and ToolFlow.}
	\label{tab:hallucination}
	\begin{tabular}{lrrr}
		\toprule
		\textbf{Hallucination Type} & \textbf{ToolWeave} & \textbf{ToolDial} & \textbf{ToolFlow} \\
		\midrule
		Tool name hallucination          & 0.0\% & 30.5\% & 5.0\% \\
		Parameter name hallucination     & 0.0\% &  3.5\% & 0.0\% \\
		Parameter value hallucination           & 20.0\% &  4.5\% & 46.5\% \\
		Assistant text output hallucination  & 0.0\% & 24.5\% & 75.5\% \\
		\midrule
		No hallucination (clean)         & \textbf{80.0\%} & 46.5\% & 16.5\% \\
		\bottomrule
	\end{tabular}
\end{table}

\subsection{LLM-as-Judge Evaluation of Dialogues}
\label{sec:llm_dialogue_eval}

Following prior work~\citep{wang2025toolflow}, we evaluate dialogue quality using an LLM-as-judge protocol. We use \texttt{GPT-5-mini} to rate 200 sampled dialogues per dataset on a 1--5 scale across four dimensions: \emph{Naturalness} (human-likeness of the conversation), \emph{Coherence} (logical flow and relevance), \emph{Helpfulness} (usefulness of the assistant responses), and \emph{Accuracy} (factual correctness). Table~\ref{tab:llm_dialogue_eval} reports the results. ToolWeave achieves the highest scores across all dimensions. In contrast, ToolFlow receives substantially lower coherence and helpfulness scores, consistent with the loop-heavy behavior observed in Table~\ref{tab:dialogue_stats}. ToolDial attains reasonable accuracy but lags in naturalness and coherence. Overall, these results confirm that ToolWeave produces dialogues that are both structurally complex and conversationally coherent.  Additional details are provided in Appendix~\ref{app:llm_judge_details}.

\begin{table}[t]
	\centering
    \small
	\caption{\small LLM evaluation of synthetic dialogues. Nat=Naturalness, Coh=Coherence, Hel=Helpfulness, Acc=Accuracy.}
	\label{tab:llm_dialogue_eval}
	\begin{tabular*}{\linewidth}{@{\extracolsep{\fill}}lcccc}
		\toprule
		\textbf{Dataset} & \textbf{Nat} & \textbf{Coh} & \textbf{Hel} & \textbf{Acc} \\
		\midrule
		ToolWeave & 4.38 & 4.12 & 3.89 & 4.38 \\
		ToolFlow & 3.95 & 2.94 & 3.21 & 3.27 \\
		ToolDial & 3.38 & 3.42 & 3.00 & 3.82 \\
		\bottomrule
	\end{tabular*}
\end{table}

%% file: sections/expt_new.tex
\section{Experiments}
\label{sec:experiments}
We evaluate ToolWeave along three questions:
\begin{enumerate}[leftmargin=*, noitemsep, topsep=0pt, partopsep=0pt]
    \item \textit{Performance Study:} How does ToolWeave compare with existing pipelines?
    \item \textit{Synthetic vs. Real APIs:} How do ToolWeave's synthetic APIs compare with real APIs?
    \item \textit{Ablation Study:} How does each component of ToolWeave affect overall performance?
\end{enumerate}

\subsection{Experimental Setup}
We synthesize ToolWeave data using \texttt{\textbf{G}PT-5-mini}, denoted as \emph{ToolWeave (G)}, and \texttt{GPT-\textbf{O}SS-120B}, denoted as \emph{ToolWeave (O)}. We use \emph{ToolWeave-R} to denote variants that employ real-world RapidAPI schemas instead of our synthetic APIs. For each configuration, we generate 3.2K dialogues. We compare against two baselines: \emph{ToolDial} (dialogues generated with GPT-4o) and \emph{ToolFlow}\footnote{We implemented ToolFlow using the prompts provided in the paper and synthesized the dialogues using \texttt{GPT-5-mini}.}. We focus on reproducible pipelines; several recent methods~\cite{yin-etal-2025-magnet, apigen-MT2025, toolAce2025iclr} lack public code or do not specify the data mixtures used for fine-tuning, making faithful implementation and fair comparison difficult.

We evaluated three base models: \emph{Llama-3.1-8B}, \emph{Llama-3.1-70B}, and \emph{Phi-4}, covering various model sizes and families. All models are fine-tuned using \texttt{QLoRA} via \texttt{LLaMA-Factory}. All hyperparameters used for fine-tuning are provided in Appendix~\ref{app:finetuning}. We evaluate all models on a diverse suite of three multi-turn tool-calling benchmarks:  \emph{BFCL-V3}~\citep{patil2025bfcl}, \emph{API Bank}~\citep{li-etal-2023-apibank}, and \emph{CONFETTI}~\citep{alkhouli-etal-2025-confetti}. Benchmark details are provided in Appendix~\ref{app:benchmark_details}.

\subsection{Performance Study}
\label{subsec:benchmark_performance}
\begin{table}[t]
		\centering
		\caption{\small  BFCL-V3 multi-turn performance.}
		\label{tab:multi_turn_results}
		\scriptsize
		\resizebox{\columnwidth}{!}{
            \begin{tabular}{lccccc}
				\toprule
				\textbf{Model} &
				\makecell{\textbf{Avg.}\\\textbf{Acc.}} &
				\textbf{Base} &
				\makecell{\textbf{Miss}\\\textbf{Func.}} &
				\makecell{\textbf{Miss}\\\textbf{Param.}} &
				\makecell{\textbf{Long}\\\textbf{Ctx.}} \\
                \midrule
				Llama-3.1-8B & 9.25 & 11.00 & 8.00 & 8.50 & 9.50 \\[3pt]
				+ ToolDial & 1.75 & 0.50 & 3.50 & 2.50 & 0.50 \\
				+ ToolFlow & 7.50 & 9.50 & 5.50 & 5.50 & 9.50 \\
				\textbf{+ ToolWeave-R (G)} & 18.38 & 24.00 & 18.00 & 14.50 & 17.00 \\
				\textbf{+ ToolWeave (O)} & 19.88 & 23.00 & \textbf{21.50} & 15.50 & 19.50 \\
				\textbf{+ ToolWeave (G)} & \textbf{21.12} & \textbf{28.00} & 18.00 & \textbf{18.50} & \textbf{20.00} \\
				\midrule
				Llama-3.1-70B & 12.50 & 17.00 & 13.00 & 10.50 & 9.50 \\[3pt]
				+ ToolDial & 3.75 & 4.50 & 3.00 & 4.00 & 3.50 \\
				+ ToolFlow & 23.50 & 29.50 & 20.50 & 20.00 & 24.00 \\
				\textbf{+ ToolWeave-R (G)} & 37.38 & 40.00 & \textbf{43.00} & 34.00 & 32.50 \\
				\textbf{+ ToolWeave (O)} & 33.25 & 37.50 & 35.00 & 32.00 & 28.50 \\
				\textbf{+ ToolWeave (G)} & \textbf{39.75} & \textbf{42.50} & \textbf{43.00} & \textbf{37.00} & \textbf{36.50} \\
				\midrule
				Phi-4 & 3.12 & 7.50 & 0.00 & 2.50 & 2.50 \\[3pt]
				+ ToolDial & 2.00 & 2.00 & 0.50 & 3.50 & 2.00 \\
				+ ToolFlow & 10.12 & 11.00 & 10.00 & 11.50 & 8.00 \\
				\textbf{+ ToolWeave-R (G)} & 26.25 & 33.50 & \textbf{33.50} & 20.50 & 17.50 \\
				\textbf{+ ToolWeave (O)} & 24.50 & {24.00} & 28.50 & 31.00 & 14.50 \\
				\textbf{+ ToolWeave (G)} & \textbf{29.50} & \textbf{36.00} & 30.50 & \textbf{31.50} & \textbf{20.00} \\
                \bottomrule
			\end{tabular}
		}
\end{table}

\begin{table}[!ht]
    \centering
    \small
    \caption{\small  BFCL-V3 multi-turn performance of \texttt{Llama-3.1-70B} across two splits: multi-step and single-step dialogues.}
		\label{tab:multi_step_split}
    \begin{tabular}{lccc}
    \toprule
         & \makecell{\textbf{Multi-step}\\\textbf{Accuracy}}& \makecell{\textbf{Single-step}\\\textbf{Accuracy}} & \makecell{\textbf{Average}\\\textbf{Accuracy}}  \\ 
        \midrule
        Base & 10.80 & 17.12 & 12.50 \\ 
        ToolFlow & 19.75 & 26.32 & 23.50 \\ 
        ToolWeave & 38.73 & 44.08 & 39.75 \\ 
        \bottomrule
    \end{tabular}
\end{table}

\begin{table}[t]
	\centering
	\caption{Generalization across API-L1, API-L2, and CONFETTI.}
	\scriptsize
	\label{tab:non-bfcl}
	\resizebox{\linewidth}{!}{%
	\begin{tabular}{lcccc}
		\toprule
		\textbf{Model} & \textbf{API-L1} & \textbf{API-L2} & \textbf{CONFETTI} & \textbf{Average} \\
		\midrule
		Llama-3.1-8B & 62.91 & 59.46 & 20.55 & 47.64 \\
		+ ToolDial & 59.15 & 52.70 & 23.12 & 44.99 \\
		+ ToolFlow & 67.92 & 54.05 & 17.39 & 46.45 \\
		+ ToolWeave-R (G) & 68.17 & 51.35 & 19.37 & 46.30 \\
		+ ToolWeave (O) & 68.92 & 60.81 & \textbf{36.76} & 55.50 \\
		+ ToolWeave (G) & \textbf{75.44} & \textbf{63.51} & 28.85 & \textbf{55.93} \\
		\midrule
		Llama-3.1-70B & 54.89 & 59.46 & 33.00 & 49.12 \\
		+ ToolDial & \textbf{71.93} & 40.54 & 11.46 & 41.31 \\
		+ ToolFlow & 70.68 & 59.46 & 28.46 & 52.87 \\
		+ ToolWeave-R (G) & 71.93 & 66.22 & 33.99 & 57.38 \\
		+ ToolWeave (O) & 71.18 & \textbf{64.86} & \textbf{45.45} & \textbf{60.50} \\
		+ ToolWeave (G) & 71.43 & 63.51 & 34.78 & 56.57 \\
		\midrule
		Phi-4 & 34.09 & 25.68 & 13.44 & 24.40 \\
		+ ToolDial & 55.64 & 44.59 & 13.64 & 37.96 \\
		+ ToolFlow & 49.12 & 56.76 & 15.61 & 40.50 \\
		+ ToolWeave-R (G) & 75.19 & 62.16 & 32.41 & 56.59 \\
		+ ToolWeave (O) & 71.68 & 62.16 & \textbf{33.60} & 55.81 \\
		+ ToolWeave (G) & \textbf{76.69} & \textbf{64.86} & 31.42 & \textbf{57.66} \\
		\bottomrule
	\end{tabular}}
\end{table}

\noindent \textbf{BFCL-v3:} Table \ref{tab:multi_turn_results} presents the results on BFCL v3 using the overall aggregated average accuracy (Avg. Acc). BFCL v3 evaluates not only the basic ability to perform tool calling (Base), but also tests whether a model can prompt the user when required information is missing (Miss Param.), refrain from hallucinating when an appropriate tool is unavailable (Miss Func), and maintain accuracy in long, information-rich contexts (Long Ctx.). The BFCL dataset contains 800 dialogues in total, of which 648 include at least one multi-step interaction, while the remaining dialogues do not contain any multi-step interactions.

When LLMs are fine-tuned on dialogues generated by ToolDial, all three models perform worse than the base model. This is due to two key limitations: a cap of two tool calls per dialogue and a reliance on only 21 fixed interaction templates. As a result, the dataset contains only 5\% multi-step examples, with predefined flows that do not align well with the more realistic interaction patterns found in BFCL. ToolFlow performs better than ToolDial because it generates more realistic dialogues. However, its high-level planner places most of the burden on the dialogue synthesizer, resulting in only 7\% multi-step dialogues. In contrast, ToolWeave produces 45\% multi-step dialogues, largely due to its fine-grained turn-level planner, thus outperforming ToolFlow by a large margin (at least 13 points). Table \ref{tab:multi_step_split} shows the performance across two subsets of the data: multi-step and single-step. Across all three settings, multi-step accuracy consistently lags single-step by roughly 5–6 percentage points. This is expected as multi-step tasks compound errors at each step. ToolWeave achieves 3.6 times the multi-step accuracy and 2.6 times the single-step accuracy of the Base model. 

\noindent \textbf{Other Benchmarks: } To assess the breadth of our approach, we evaluate ToolWeave across two additional benchmarks. Table~\ref{tab:non-bfcl} shows \emph{ToolWeave} consistently improves performance across benchmarks, with average gains reaching 33.26 points (Phi-4). Notably, our 16.2 point improvement on CONFETTI demonstrates superior handling of complex conversational dynamics such as goal switching and chained function calls, where performance is typically limited across state of the art models. These results demonstrate that ToolWeave enhances multi-step reasoning and compositional tool use.

\subsection{Synthetic vs.\ Real APIs}
We further compare ToolWeave’s synthetic APIs against real APIs from RapidAPI. As shown in Table~\ref{tab:multi_turn_results}, \texttt{Llama-3.1-70B} achieves 39.75\% accuracy when trained on fully synthetic ToolWeave data, compared with 37.38\% when trained on RapidAPI-based data (ToolWeave-R). One possible explanation is that real-world APIs often lack explicit parameter dependency structure, which can introduce noisy links in the tool graph. In contrast, ToolWeave’s synthetic tool graphs explicitly encode parameter relationships, producing a cleaner training signal for multi-step tool usage.

\subsection{Ablation Study}
\label{sec:ablations}

\begin{table}[h]
	\centering
	\caption{Component ablation on \texttt{Llama-3.1-8B-Instruct} (BFCL-V3).}
	\label{tab:component_ablation}
	\resizebox{\columnwidth}{!}{
		\begin{tabular}{lc}
			\toprule
			\textbf{Setting} & \textbf{Average Acc.} \\
			\midrule
			\textbf{Full ToolWeave Pipeline} & \textbf{19.88} \\
			\quad w/o Post-Processing & 13.62 \\
			\quad w/o Structured Sampler (Random Walk) & 12.25 \\
			\quad w/o Fine-Grained Planner (ToolFlow-style) & 7.50 \\
			\bottomrule
		\end{tabular}
	}
\end{table}

We perform ablation experiments on BFCL v3 using GPT-OSS-120B due to budget constraints. Table~\ref{tab:component_ablation} reveals that each module in ToolWeave is critical for realistic multi-step tool-calling dialogue synthesis. Removing the fine-grained turn-level planner in favor of a high-level ToolFlow-style planner causes performance to collapse to 7.50\%, confirming that models cannot synthesize complex trajectories without turn-level planning. Replacing our tool graph sampler with a random walk reduces accuracy to 12.25\%. Removing post-processing leads to a 6.26\% drop, validating the necessity of simulating real-world errors and recovery behaviors. Detailed breakdown of metrics for these ablations is provided in Appendix~\ref{app:component_ablations}.

%% file: sections/conclusion.tex
\section{Conclusion}
\label{sec:conclusion}
We introduced ToolWeave, a modular framework for synthesizing realistic multi-turn tool-calling datasets through structured generation stages that enforce multi-step interactions and consistent parameter provenance. Across multiple LLM families and benchmarks, models fine-tuned on ToolWeave data significantly outperform existing pipelines. Our results reveal that structured synthesis enables open-weight models to produce higher-quality training data than frontier-model-driven pipelines. We further show that ToolWeave's synthetic APIs can rival or surpass real-world APIs such as those from RapidAPI. Overall, our findings suggest that the structure of the synthesis pipeline, rather than the scale of the synthesis LLM, is the key driver of realistic multi-turn tool-calling data.

%% file: sections/appendix_two_column.tex
\section{Comparison of ToolWeave with other SDG pipelines}
\label{sec:sdg_pipeline_comparsion} 
Table~\ref{tab:sdg-pipeline-comparison} provides the comparison of the features around licenses of the model used, the code, the generated data, and the fine-grained control it offers.
\label{sec:relWorkTable}
\begin{table*}[htb]
	\centering
	\resizebox{1.0\linewidth}{!}{%
		\begin{threeparttable}
			\caption{Comparison of multi-turn synthetic data pipelines across different dimensions.}
			\label{tab:sdg-pipeline-comparison}
			\begin{tabular}{lccccc}
				\toprule
				\textbf{Framework} &
				\makecell{\textbf{Synthesize}\\\textbf{APIs}} &
				\makecell{\textbf{Code}\\\textbf{d}} &
				\makecell{\textbf{Data Released,}\\\textbf{License Friendly}} &
				\makecell{\textbf{License Friendly}\\\textbf{Synthesis Model}} &
				\makecell{\textbf{Fine-Grained}\\\textbf{Plan}} \\
				\midrule
				Tooldial~(\citeauthor{shim2025tooldial})   & \xmark & \cmark & \cmark, \xmark & \xmark & \xmark \\
				ToolAce~(\citeauthor{toolAce2025iclr})    & \cmark & \xmark & \cmark, \xmark & \xmark & \xmark \\
				APIgen-MT~(\citeauthor{apigen-MT2025})  & \xmark & \xmark & \cmark, \xmark & \xmark & \xmark \\
				ToolFlow~(\citeauthor{wang2025toolflow})   & \xmark & \xmark & \xmark, \xmark & \xmark & \xmark \\
				Button~(\citeauthor{button-2025-iclr})     & \cmark & \xmark & \cmark, \xmark & \xmark & \xmark \\
				ToolLLM~(\citeauthor{qin2024toolllm})    & \xmark & \cmark & \cmark, \xmark & \xmark & \xmark \\
				ToolWeave (Ours)  & \cmark & \cmark& \cmark, \cmark & \cmark & \cmark \\
				\bottomrule
			\end{tabular}
		\end{threeparttable}%
	}
\end{table*}

\section{Tool Graph Synthesizer Details}
\label{app:synthethic_apis_details}
This section provides additional implementation details of the Tool Graph Synthesizer, including the schema validation procedures, and graph construction strategies.

\subsection{Domains for Tool Graph Synthesis}
\label{app:domain_list}

The 20 domains used to generate the synthetic API pool are:

\begin{tcolorbox}[colback=white, colframe=black, boxrule=1pt, sharp corners, width=\columnwidth, left=2pt, right=2pt, boxsep=2pt]
	\small
	\begin{multicols}{2}
		\setlength{\columnsep}{10pt}
		\begin{itemize}[noitemsep, topsep=0pt, parsep=0pt, leftmargin=*, label={}]  
			\item Agriculture
			\item Customer Support
			\item Cybersecurity
			\item E-commerce
			\item Education
			\item Energy
			\item Film Industry
			\item Human Resources
			\item Insurance
			\item Legal Services
			\item Logistics
			\item Manufacturing
			\item Marketing
			\item Online Banking
			\item Real Estate
			\item Retail
			\item Supply Chain
			\item Telecommunications
			\item Tourism
			\item Transportation
		\end{itemize}
	\end{multicols}
\end{tcolorbox}

\subsection{Sample Tool Graph}
\label{app:sample_tool_graph}

\begin{figure*}[t]
		\centering
		\includegraphics[width=0.8\linewidth]{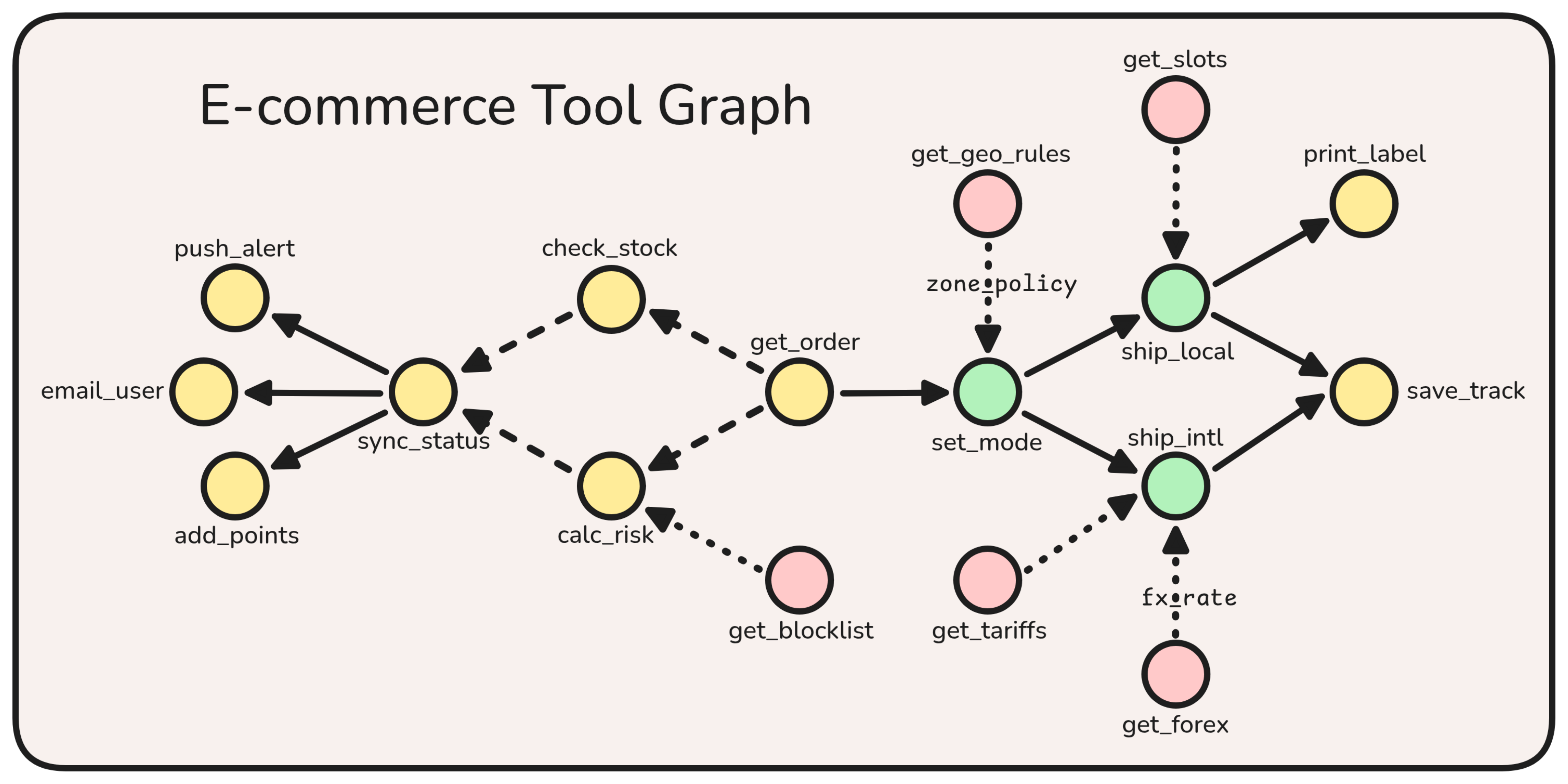}
		\caption{E-commerce tool graph demonstrating linear, fan-in-fan-out (dashed), and conditional (green) dependency patterns. Red nodes indicate auxiliary tools providing specific configuration parameters, such as foreign exchange rates or zone policies, to the primary workflow.}
		\label{fig:ecommerce_tool_graph}
\end{figure*}

Figure \ref{fig:ecommerce_tool_graph} illustrates a representative tool graph for the E-commerce domain. The nodes represent discrete tools (APIs), connected by directed edges that signify data dependencies; specifically, a parameter from the output schema of a predecessor node maps to an input parameter of the successor (e.g., \texttt{zone\_policy} from \texttt{get\_geo\_rules} acts as an input for \texttt{set\_mode}).

The graph topology highlights three distinct execution patterns, corresponding to the Tool-Graph traversal strategies discussed in Section \ref{sec:goal_generation}:
\begin{itemize}
    \item \textbf{Linear Chains:} Standard sequential dependencies connected by any type of edge, such as \texttt{get\_order} $\rightarrow$ \texttt{set\_mode} $\rightarrow$ \texttt{ship\_local} $\rightarrow$ \texttt{save\_track}.
    \item \textbf{Fan-in-Fan-out (Dashed):} Illustrated by the sequence \texttt{get\_order} $\rightarrow$ \{\texttt{check\_stock}, \texttt{calc\_risk}\} $\rightarrow$ \texttt{sync\_status}. Here, the initial tool provides inputs to a parallel stage of tools that are independent of one another, the outputs of which collectively feed into a common successor.
    \item \textbf{Conditional Branching (Green):} Represented by the \texttt{set\_mode} node, where the downstream execution path (\texttt{ship\_local} vs. \texttt{ship\_intl}) is determined dynamically based on the output value of the decision node.
\end{itemize}

\subsection{Detailed Algorithms for Tool Graph Synthesizer}

This section expands upon the Tool Graph Synthesizer process from Section~\ref{sec:tool_graph_synthesizer}. While the description in the main paper presents the high-level methodology, here we provide the main algorithm, the synthesis plan, and detailed sub-algorithms for key steps like domain context construction and API refinement.\\

\begin{algorithm}[t]
\footnotesize
	\caption{Tool Graph Synthesis}
	\label{alg:api_synth_main}
	\KwIn{Domain descriptor $D$, synthesis plan $P$, LLM $L$, embedding model $E$}
	\KwOut{API pool $\mathcal{A}$, Tool Graph $\mathcal{G}$}
	Initialize $\mathcal{A} \leftarrow \emptyset$, $\mathcal{G} \leftarrow \emptyset$\;
	$C \leftarrow \textsc{ConstructDomainContext}(D)$  \tcp*{See Appendix~\ref{app:algo_domain_context}}
	\ForEach{step $s \in P$}{
		Generate $n_s$ candidate APIs with $L$ using domain context $C$ and prompt from step $s$\;
		Parse and validate candidates syntactically\;
		Deduplicate candidates against $\mathcal{A}$ using embeddings from $E$ and structural checks\;
		Refine and semantically validate schemas (defaults, enums, nested objects, etc.)\;
		Add accepted APIs to $\mathcal{A}$\;
	}
	Construct Tool Graph $\mathcal{G}_{tool}$ by validating data-flows between APIs in $\mathcal{A}$\;
	\Return{$\mathcal{A}, \mathcal{G}_{tool}$}
\end{algorithm}

\subsection{Synthesis Plan}
\label{app:synthesis_plan}

Our synthesis plan acts as a curriculum for the API-generating LLM. It specifies a sequence of generation steps, each with a unique prompt template and a target number of APIs. This design is extensible and allows us to control the breadth and depth of the final API pool. Our implementation consists of five main stages:

\begin{enumerate}[leftmargin=*,noitemsep,topsep=0pt,parsep=0pt]
	\item \textbf{Seed Generation:} This initial stage generates a small batch of fundamental, ``entry-point'' APIs for the domain (e.g., \texttt{search\_product} in e-commerce).
	
	\item \textbf{Entity Expansion:} This stage uses the entities extracted from Wikidata (see Appendix~\ref{app:algo_domain_context}) to generate new APIs that specifically cover domain-relevant entities, ensuring breadth.
	
	\item \textbf{Schema Enrichment:} This stage focuses on depth. It takes existing, simple APIs and iteratively increases their structural complexity by adding nested objects, enums, default values, and required flags to mimic real-world enterprise-grade schemas.
	
	\item \textbf{Connection Discovery:} It generates new APIs that plausibly connect existing APIs, creating data-flow paths (e.g., a \texttt{get\_product\_details} API that takes a \texttt{product\_id} from \texttt{search\_product}).
	
	\item \textbf{Pattern Expansion:} This final stage diversifies the API pool by generating parallel variations of existing APIs (e.g., different ways to search, like \texttt{search\_by\_name}, \texttt{search\_by\_category}, etc.).
\end{enumerate}
For reproducibility, we include the  synthesis plan used in our experiments below. Prompt paths have been abstracted for clarity.
\begin{figure*}
\begin{minted}[
  frame=double,
  fontsize=\footnotesize,
  breaklines,
  breakanywhere,
  autogobble,
  xleftmargin=0.5em,
  framesep=2mm
]{json}
{
    "steps": [
        {"name": "Seed Generation", "num_to_generate": 8},
        {"name": "Entity Expansion", "num_to_generate": 8},
        {"name": "Schema Enrichment", "num_to_generate": 5},
        {"name": "Connection Discovery", "num_to_generate": 8},
        {"name": "Pattern Expansion", "num_to_generate": 5}
    ]
}
\end{minted}
\end{figure*}
\subsection{Detailed Sub-Algorithms}
The main generation algorithm (Algorithm~\ref{alg:api_synth_main}) relies on several key sub-routines, which are detailed below.

\subsubsection{Domain Context Construction}
\label{app:algo_domain_context}
To provide domain grounding for API synthesis, we construct a domain context using external knowledge sources (Algorithm~\ref{alg:create_domain_context}). Specifically, we retrieve a Wikipedia summary and structured metadata from Wikidata, including entity descriptions, subclasses, and related properties. These signals are consolidated into a compact prompt representation used during generation. The resulting context constrains the synthesis process to domain-relevant entities, operations, and relationships.

\begin{algorithm}[t]
\footnotesize
	\caption{CreateDomainContext}
    \label{alg:create_domain_context}
	\KwIn{Domain name string $d$}
	\KwOut{Domain context $C$}
	Resolve $d$ to canonical Wikidata entity $QID$\;
	Retrieve Wikipedia summary $T_{wiki}$ for $d$\;
	Extract structured facts $F_{wikidata}$ (classes, subclasses, properties) from $QID$\;
	Assemble $C \leftarrow (T_{wiki}, F_{wikidata})$\;
	\Return{$C$}
\end{algorithm}

\subsubsection{API Refinement}
\label{app:algo_refine}
Generated APIs are post-processed to improve schema consistency and usability (Algorithm~\ref{alg:refine_api}). This stage removes redundant or echo parameters, resolves required fields, and refines parameter properties such as enums, default values, and formats. Duplicate APIs are filtered using lexical, structural, and semantic similarity checks. These steps produce a consistent and non-redundant API pool for subsequent workflow construction.

\begin{algorithm}[t]
\footnotesize
	\caption{RefineAPI}
    \label{alg:refine_api}
	\KwIn{Candidate API $a$, current API set $\mathcal{A}$}
	\KwOut{Refined API $a'$ or $\emptyset$}
	Check if $a$ is duplicate of any in $\mathcal{A}$ (lexical, structural, semantic)\;
	\If{duplicate}{
		\Return{$\emptyset$}
	}
	Remove echo parameters (outputs attributes that are same as inputs)\;
	Enrich schema: infer enums, default values, required fields\;
	Normalize parameter names and types\;
	Paraphrase descriptions for stylistic variance\;
	\Return{refined API $a'$}
\end{algorithm}

\subsubsection{Connection Graph Construction}
\begin{algorithm}[t]
\footnotesize
	\caption{ConstructToolGraph}
    \label{alg:construct_toolgraph}
	\KwIn{API set $\mathcal{A}$}
	\KwOut{Connection graph $\mathcal{G}$}
	Initialize graph $\mathcal{G} = (V=\mathcal{A}, E=\emptyset)$\;
	\ForEach{ordered pair $(a_i, a_j) \in \mathcal{A} \times \mathcal{A}$}{
		Identify candidate parameter matches $(p_{out}, p_{in})$ \tcp*{By param name matches}
		Validate match using LLM judgment\;
		\If{valid}{
			Add edge $(a_i.p_{out} \rightarrow a_j.p_{in})$ to $E$\;
		}
	}
	\Return{$\mathcal{G}$}
\end{algorithm}

Matching parameters between APIs by exact name is appropriate in our setting because the API synthesis prompts explicitly instruct the model to create connections to previously generated APIs only through parameters with identical names (see Table~\ref{alg:construct_toolgraph}). This design encourages consistent parameter naming and simplifies the construction of tool dependencies.

\subsection{API Quality Metric Definitions}
\label{app:api_metrics}

This section provides the formal definitions for the metrics used to evaluate the quality of the synthetic APIs in Section~\ref{sec:api_quality}. In all definitions, $A = \{a_1, a_2, \dots, a_n\}$ represents the set of $n$ APIs in a given domain.

\subsubsection*{1. Interconnectivity $(IC)$}
The average number of input parameters per API that can be filled by an output from another API in the graph. This measures the data-flow potential.

\[
IC = \frac{\sum_{i=1}^{n}\left|I_{a_i} \cap O\right|}{n}
\]

\noindent
Where:
\begin{itemize}[noitemsep,topsep=0pt,parsep=0pt]
	\item $O = \bigcup_{i=1}^{n} O_{a_i}$: The set of all unique output parameters from all APIs.
	\item $I_{a_i}$: The set of input parameters for API $a_i$.
	\item $|I_{a_i} \cap O|$: The count of input parameters for $a_i$ that match any output from the entire API set $A$.
\end{itemize}

\subsubsection*{2. Complex API Use $(CAU)$}
The proportion of APIs that use at least one complex parameter type (i.e., ``object'' or ``array''), measuring schema depth.

\[
CAU = \frac{\sum_{i=1}^{n} C(a_i)}{n}
\]

\noindent
Where the indicator function $C(a_i)$ is defined as:

\begin{equation*}
	C(a_i)=
	\begin{cases}
		1, & \parbox[t]{0.6\columnwidth}{if $a_i$ has an \texttt{object} or \texttt{array} parameter},\\
		0, & \text{otherwise}
	\end{cases}
\end{equation*}

\subsubsection*{3. Required Parameter Ratio (RPR)}
The average proportion of an API's input parameters that are marked as ``required'', measuring schema strictness.

\[
RPR = \frac{1}{n} \sum_{i=1}^{n} \frac{|I_{\text{req}}(a_i)|}{|I_{\text{total}}(a_i)|}
\]
\noindent
Where $I_{\text{req}}(a_i)$ is the set of required input parameters for API $a_i$, and $I_{\text{total}}(a_i)$ is the set of all input parameters for $a_i$.

\subsubsection*{4. Longest Chain Length}
The length of the longest simple path (a path with no repeated nodes) in the \emph{Tool Graph}. This metric quantifies the maximum number of sequential, multi-step operations possible in a single dialogue.

\section{Goal Generation Implementation Details}
\label{app:goal_gen_details}

This section provides the low-level implementation details for the \emph{Goal Generation} stage (Section~\ref{sec:goal_generation}), including the algorithms and heuristics used to extract, score, and synthesize goals. 

\subsection{Pattern Extraction Algorithms}
The main paper outlines three pattern classes. Here is how they are discovered:

\begin{enumerate}[leftmargin=*,noitemsep,topsep=0pt,parsep=0pt]
	\item \textbf{Linear Paths:} These are found using a bounded \emph{beam search}~\citep{norvig2002modern}. The search explores paths from a start node, expanding a ``beam'' of the top-$k$ candidates at each depth.
	
	\item \textbf{Fan-in / Fan-out:} These patterns are found by analyzing node connectivity. For a given \texttt{start\_node}, we find its successors. We then find the ``common children'' of combinations of these successors using set intersection. This allows us to identify points where parallel data flows (fan-out) later merge (fan-in).
	
	\item \textbf{Conditional Branches:} These are identified by scanning an API's output schema. We scan for output fields that can act as a logical predicate, specifically those typed as \texttt{boolean}, \texttt{enum}, or other simple types, while filtering out non-conditional fields like IDs.
\end{enumerate}

\subsection{Path Scoring and Ranking}
To rank the paths found by our algorithms, we use a combination of a hybrid score and a diversification algorithm.

\paragraph{Hybrid Heuristic Score:}

The \texttt{final\_score} for any given path is a weighted sum of three components, and is calculated as:

\begin{equation*}
\begin{aligned}
S_{\text{final}} ={}& w_l \cdot (S_{\text{coherence}} + S_{\text{relevance}}) \\
&+ w_d \cdot S_{\text{dataflow}} + w_b \cdot S_{\text{length}}
\end{aligned}
\end{equation*}

Where:
\begin{itemize}[leftmargin=*,noitemsep,topsep=1.0pt,parsep=1pt]
	\item $S_{\text{coherence}}$ and $S_{\text{relevance}}$ are \emph{LLM-based ratings} (from $-2$ to $+2$) where the LLM judges the quality of a synthesized goal for the provided tool path.
	\item $S_{\text{dataflow}}$ is the \emph{semantic dataflow score}, calculated by checking cosine similarity between embeddings of output and input parameters of sequential tools.
	\item $S_{\text{length}}$ is a \emph{length bonus} to reward longer, more complex paths.
	\item $w_l$, $w_d$, and $w_b$ are the respective weights for LLM rating, dataflow, and length, set as hyperparameters (we use $w_l=0.5, w_d=0.8, w_b=0.3$ in our experiments).
\end{itemize}

\paragraph{Ranking and Embedding Model:}
To ensure a diverse set of goals, we rank the top-scoring linear paths using \emph{Maximal Marginal Relevance (MMR)}~\citep{carbonell1999mmr}. This balances the \texttt{final\_score} (relevance) with the cosine dissimilarity from already-selected goals (diversity). In our experiments, we use a value of $0.7$ for $\lambda$. All embeddings for MMR and semantic dataflow are computed using the \texttt{all-MiniLM-L6-v2} model from the \emph{SentenceTransformers}~\citep{reimers2019sentence-bert} library.

\subsection{Goal Generation Algorithm}\label{app:algo_goal_generation}
The goal generation algorithm~\ref{alg:goal_generation_domain} converts sampled tool paths into natural-language task descriptions. Given a sequence of tools, the algorithm constructs a goal that reflects the functional intent of the workflow while preserving dependencies between tool inputs and outputs. Candidate goals are generated and filtered to ensure semantic coherence with the underlying tool sequence. This procedure associates each executable tool path with a clear user-oriented objective.
\begin{algorithm*}[t]
    \renewcommand{\baselinestretch}{1.25}\selectfont
	\caption{Domain-level Goal Generation}
	\label{alg:goal_generation_domain}
	\KwIn{Tool graph for domain $\mathcal{G}_{tool}$, beam width $B$, max depth $D_{\max}$, selection size $K$, MMR weight $\lambda$, LLM $L$, embedding model $E$}
	\KwOut{Goal set $\mathcal{G}_{goals}$ (JSON objects: \texttt{\{tool\_path, pattern\_type, goal\_text, metadata\}})}
	Initialize $\mathcal{G}_{goals} \leftarrow \emptyset, \mathcal{S} \leftarrow \emptyset,$ \,where $\mathcal{S}$ is the set of selected tool paths\;
	$S_{linear} \leftarrow \textsc{BeamSearch}(\mathcal{G}_{tool}, B, D_{\max}, L)$\;
	$S_{linear} \leftarrow \textsc{TopKMMR}(S_{linear}, K, \lambda, E)$\;
	
	$S_{fan} \leftarrow \textsc{FindFanInFanOut}(\mathcal{G}_{tool})$\;
	
	$S_{cond} \leftarrow \textsc{FindConditional}(\mathcal{G}_{tool})$\;
	
	$S \leftarrow S_{linear} \cup S_{fan} \cup S_{cond}$\;
	
	\ForEach{$p \in S$}{
		$prompt \leftarrow \textsc{BuildGoalPrompt}(p,\; \textsc{PatternType}(p))$\;
		$goal\_text \leftarrow L.\textsc{Generate}(prompt)$\;
        $obj \leftarrow \textsc{BuildJSON}\big(p,\; \textsc{PatternType}(p),\; goal\_text,\; \textsc{Metadata}(p)\big)$\;
		$\mathcal{G}_{goals} \leftarrow \mathcal{G}_{goals} \cup \{obj\}$\;
	}
	\Return{$\mathcal{G}_{goals}$}
\end{algorithm*}

\subsection{Example Goal Synthesis Prompt}
The final step, `Goal synthesis via LLM', uses different prompts for each pattern type. The prompt provides the LLM with the tool sequence and their schemas, and asks it to generate a natural-language goal.


\section{Dialogue Planner Implementation Details}
\label{app:dialogue_planner}
This section provides the low-level implementation details for the \emph{Dialogue Planner} stage (Section~\ref{sec:dialogue_planner}), including the algorithm and the logic used to partition tool paths, generate subgoals, plan parameters, and inject interaction patterns. 

\subsection{Dialogue Planner Algorithm}
Algorithm~\ref{alg:dialogue_planner} summarizes the high-level process, highlighting subgoal generation, speaker sequencing, parameter planning, and clarification insertion.

{
    \renewcommand{\baselinestretch}{1.25}\selectfont
    \begin{algorithm*}[t]
    	\caption{Dialogue Planning}
    	\label{alg:dialogue_planner}
    	\KwIn{Tool path $P$, pattern type $T \in \{\texttt{linear},\texttt{fan},\texttt{conditional}\}$, overall goal text $G$, relevant tool schemas $\Sigma$, goal metadata $M$, clarification probability $p_{clar}$, LLM for partitioning and subgoal synthesis $L$}
    	\KwOut{Ordered plan $\mathcal{D}$ (step-level JSON objects with keys: \texttt{step\_idx}, \texttt{role}, \texttt{subgoal}, \texttt{tools}, \texttt{params}, \texttt{metadata})}
    	
    	Initialize partition list $\mathcal{P}^* \leftarrow \textsc{PartitionToolPath}(P, T, \Sigma, L)$, subgoal text list $\mathcal{S} \leftarrow [\ ]$ \;
    	\ForEach{partition $p \in \mathcal{P}^*$}{
    		$\mathcal{S}.\textsc{append}(\textsc{SynthesizeSubgoal}(p, G, L))$ \;
    	}
    	
    	$\mathcal{D} \leftarrow [\ ]$ \;
    	\For{$i \leftarrow 1$ \KwTo $|\mathcal{P}^*|$}{
    		$p \leftarrow \mathcal{P}^*[i]$\,, $s \leftarrow \mathcal{S}[i]$ \;
    		$\mathsf{Params} \leftarrow \textsc{ExtractToolParams}(p, \Sigma)$ \;
    		$\mathsf{Params} \leftarrow \mathsf{Params} \setminus \textsc{UpstreamToolOutputs}(\mathcal{P}^*, i-1, \Sigma)$ \;
    		$(\mathsf{ReqParams}, \mathsf{ClarParams}) \leftarrow \textsc{SplitParams}(\mathsf{Params}, p_{clar})$ \;
    		
    		$\mathcal{D}.\textsc{append}(\textsc{CreateUserTurn}(i, s, p, \mathsf{ReqParams}))$ \;
    		
    		\If{$\mathsf{ClarParams} \neq \emptyset$}{
    			$\mathcal{D}.\textsc{append}(\textsc{CreateAssistantClarificationTurn}(i, \mathsf{ClarParams}))$ \;
    			$\mathcal{D}.\textsc{append}(\textsc{CreateUserClarificationTurn}(i, \mathsf{ClarParams}))$ \;
    		}
    		
    		\ForEach{tool $t \in p$}{
    			$\mathsf{ToolParams} \leftarrow \textsc{ExtractToolParams}([t], \Sigma)$ \;
    			$\mathcal{D}.\textsc{append}(\textsc{CreateAssistantToolCallTurn}(i, t, \mathsf{ToolParams}, M))$ \tcp*{Goal metadata $M$ attaches decision variables for conditionals}
    		}
    		
    		$\mathcal{D}.\textsc{append}(\textsc{CreateAssistantSummaryTurn}(i, p))$ \;
    	}
    	\Return{$\mathcal{D}$}
    \end{algorithm*}
}

\subsection{Tool Path Partitioning}
The \textsc{PartitionToolPath} function uses different strategies based on the pattern type:
\begin{itemize}[leftmargin=*,noitemsep,topsep=0pt,parsep=0pt]
	\item \textbf{Linear:} We use an LLM-based approach. The prompt provides the full tool path and their schemas, and asks the LLM to split them into coherent segments, based on logical groupings and dataflow, while also ensuring meaningful multi-step tool sequences.
	\item \textbf{Fan-in / Fan-out:} We use a deterministic graph-based approach. We identify the branching point (fan-out) and the merging point (fan-in) in the path, and create partitions accordingly. The intermediate tools between these points are grouped into either segment, and all possible valid partitions are generated to allow for diverse dialogue structures.
	\item \textbf{Conditional:} Based on the value of the decision variable (from goal metadata $M$), the second tool is chosen from the available branches. The first tool and this chosen branch form a single partition to ensure the conditional logic is preserved in the dialogue, while also generating high-quality multi-step dialogues.
\end{itemize}

\subsection{Subgoal Synthesis}
The \textsc{SynthesizeSubgoal} function uses an LLM prompt that provides the tool schemas in the partition and the overall goal text $G$. The LLM is instructed to generate a concise, user-friendly subgoal that accurately reflects the purpose of the tools in the partition, while ensuring it is distinct from other subgoals in the dialogue.

\subsection{Parameter Planning and Clarification Insertion}
The parameter planning and clarification insertion process involves several steps to ensure that the dialogue captures realistic user-assistant interactions.

\subsubsection{Parameter Preparation}
The \textsc{ExtractToolParams} function retrieves all input parameters for the tools in the current partition. The \textsc{UpstreamToolOutputs} function identifies any parameters that can be automatically filled from outputs of tools in previous partitions, and these are removed from the required parameters.

\subsubsection{Clarification Insertion}
The \textsc{SplitParams} function randomly selects a subset of the remaining parameters to be clarified, based on the clarification probability $p_{clar}$. This introduces variability and simulates real-world scenarios where users may not provide all necessary information upfront.

\subsubsection{Speaker Sequencing and Turn Creation}
The turn creation functions add well-formatted JSON objects to the dialogue plan, specifying the role and the associated content (subgoal, tool calls, parameter requests, clarifications, summaries). Following is the structure of these turns:
\begin{itemize}[leftmargin=*,noitemsep,topsep=0pt,parsep=0pt]
	\item \texttt{CreateUserTurn}: User turn with subgoal and parameters to provide upfront.
	\item \texttt{CreateAssistantClarificationTurn}: Assistant turn asking for clarification on missing parameters required for the tool calls in the current partition.
	\item \texttt{CreateUserClarificationTurn}: User turn providing values for the requested parameters.
	\item \texttt{CreateAssistantToolCallTurn}: Assistant turn invoking a tool with the provided parameters and source hints for each parameter, either from the user or from upstream tool outputs.
	\item \texttt{CreateAssistantSummaryTurn}: Assistant turn summarizing the results of the tools in the current partition.
\end{itemize}

\section{Dialogue Synthesizer Implementation Details}\label{app:dialogue_synthesis}
This section provides the low-level implementation details for the \emph{Dialogue Synthesizer} stage (Section~\ref{sec:dialogue_synthesizer}), including the overall algorithm and the logic used to build context for each role across agents. 

\subsection{Dialogue Synthesis Algorithm}
Algorithm~\ref{alg:dialogue_synthesis} summarizes the high-level process, highlighting context construction, dynamic role-based agent selection, response generation, and memory updates.

\begin{algorithm*}[h]
	\caption{Plan-Driven Dialogue Synthesis}
	\label{alg:dialogue_synthesis}
	\KwIn{Ordered plan $\mathcal{D}$, tool schemas $\Sigma$, language models for agents $L$}
	\KwOut{Dialogue transcript $\mathcal{T}$}
	Initialize memory $\mathcal{M} \leftarrow \emptyset$, transcript $\mathcal{T} \leftarrow [\,]$\;
	\ForEach{step $d \in \mathcal{D}$}{
		$ctx \leftarrow \textsc{BuildContext}(\mathcal{M}, d, \Sigma)$ \tcp*{Get memory and plan step}
		$role \leftarrow d.\texttt{role}$\;
		$agent\_prompt \leftarrow \textsc{SelectPrompt}(role)$ \tcp*{Prompt with agent's role}
		
		$response \leftarrow L.\textsc{Generate}(agent\_prompt, ctx)$ \tcp*{Utterance/tool call}
		
		$\mathcal{T}.\textsc{append}(response)$\;
		$\mathcal{M} \leftarrow \textsc{UpdateMemory}(\mathcal{M}, response, d)$ \tcp*{Update memory with output}
	}
	\Return{$\mathcal{T}$}
\end{algorithm*}


\section{Dialogue Post-Processing}
\label{app:robustness_algos}

\subsection{Dialogue Error Injection}
\label{app:dialouge_error_injector}
To rigorously evaluate and improve the robustness of models trained on ToolWeave data, we implemented a systematic error injection pipeline. This pipeline deterministically introduces schema violations, logical ordering errors, and tool confusion scenarios to force the model to learn error recovery in simulated real-world settings. Algorithm~\ref{alg:main_injection} details the high-level injection logic, while Algorithm~\ref{alg:cascading} illustrates the logic required to synthesize cascading failure scenarios.

\begin{algorithm*}[ht]
	\caption{Dialogue Error Injection}
	\label{alg:main_injection}
	\KwIn{Dialogue set $\mathcal{D}$, Tool Schemas $\mathcal{S}$, Injection Probability $p_{inject}$, Similarity Matcher $M$}
	\KwOut{Augmented Dialogue set $\mathcal{D}_{aug}$}
	
	Initialize $\mathcal{D}_{aug} \leftarrow \emptyset$\;
	\ForEach{dialogue $d \in \mathcal{D}$}{
		Add original $d$ to $\mathcal{D}_{aug}$\;
		
		\If{$Random(0, 1) < p_{inject}$}{
			$d_{error} \leftarrow \text{DeepCopy}(d)$\;
			$injected \leftarrow \text{False}$\;
			
			\tcp{Attempt Complex Logical Errors (30\% chance)}
			\If{$Random(0, 1) < 0.3$}{
				\If{\textsc{InjectCascadingFailure}($d_{error}$, $\mathcal{S}$) is Success}{
					$injected \leftarrow \text{True}$\;
				}
				\ElseIf{\textsc{InjectOutOfOrderError}($d_{error}$, $\mathcal{S}$) is Success}{
					$injected \leftarrow \text{True}$\;
				}
				\ElseIf{\textsc{InjectWrongToolError}($d_{error}$, $\mathcal{S}$, $M$) is Success}{
					$injected \leftarrow \text{True}$\;
				}
			}
			\tcp{Attempt Simple Parameter Errors (70\% chance or if complex failed)}
			\If{not $injected$}{
				\textsc{InjectSchemaError}($d_{error}$, $\mathcal{S}$)\;
			}
			
			\If{error injected successfully}{
				Mark $d_{error}$ as modified\;
				Add $d_{error}$ to $\mathcal{D}_{aug}$\;
			}
		}
	}
	\Return{$\mathcal{D}_{aug}$}
\end{algorithm*}

\begin{algorithm*}[ht]
	\caption{Inject Cascading Failure}
	\label{alg:cascading}
	\KwIn{Dialogue $d$, Tool Schemas $\mathcal{S}$}
	\KwOut{Modified Dialogue $d'$, Success boolean}
	
	Identify long multi-step sequences $T_{seq}$ (len $\ge 3$) in $d$\;
	\If{$T_{seq}$ is empty}{ \Return{$d$, False} }
	Select random sequence $S \in T_{seq}$\;
	
	\tcp{1. Simulate Reverse-Order Execution}
	Initialize failure\_turns $\leftarrow \emptyset$\;
	\For{$i \leftarrow$ length($S$)-1 \textbf{down to} 1}{
		$step_{curr} \leftarrow S[i]$, $step_{prev} \leftarrow S[i-1]$\;
		Identify param $p_{dep}$ in $step_{curr}$ dependent on $step_{prev}$\;
		
		Create turn $t_{call}$: Call $step_{curr}$ \textbf{missing} $p_{dep}$\;
		Create turn $t_{err}$: Error "Prerequisite step not completed/Missing input"\;
		Append $(t_{call}, t_{err})$ to failure\_turns\;
	}
	
	\tcp{2. Inject Full Recovery}
	Construct new history: $history_{original} + failure_{turns} + S_{original}$\;
	Update $d'.conversations$ and re-index $d'.plan$ with error steps\;
	\Return{$d'$, True}
\end{algorithm*}

\paragraph{Additional Error Injection Modules:}
Beyond the cascading failures detailed in Algorithm~\ref{alg:cascading}, our pipeline includes:

\begin{itemize}[leftmargin=*,noitemsep,topsep=0pt]
	\item \textbf{Out-of-Order Execution:} We simulate scenarios where the model attempts to execute a tool call before its dependencies are met. Specifically, we identify dependent tool pairs and inject a premature call to the downstream tool (missing the dependent parameter) prior to the prerequisite tool. The model receives a missing dependency error and must recover by executing the sequence in the correct order.
	
	\item \textbf{Semantic Confusion (Wrong Tool):} We utilize a hybrid similarity matcher (SentenceBERT~\citep{reimers2019sentence-bert} + Levenshtein distance~\citep{lcvenshtcin1966binary}) to identify semantically similar but incorrect tools (e.g., \texttt{search\_ticket} vs. \texttt{get\_ticket\_details}). We inject a turn where the model mistakenly calls the confusable tool, receives an unhelpful response, and must self-correct.
	
	\item \textbf{Schema Violations:} We deterministically mutate valid tool calls to violate JSON schema constraints. This includes removing required parameters (\texttt{MISSING\_PARAM}), violating type constraints (e.g., passing an integer for a string field), or using invalid enum values. The model is trained to read the resulting standardized error message and retry with the correct schema.

    \item \textbf{Missing Function Recovery:} We simulate scenarios where a necessary tool is initially absent from the provided context. The algorithm identifies a scheduled tool call in the plan and temporarily ``hides'' it. We inject an Assistant turn refusing the request due to the missing tool, followed by a User turn that explicitly provides the missing tool's schema JSON. The model is trained to recognize the new tool definition in the context window and immediately proceed with the correct tool call.
    
\end{itemize}


\section{Fine-tuning Hyperparameters}
\label{app:finetuning}

We fine-tune all models on 8 NVIDIA A100 SXM4 80GB GPUs using \texttt{QLoRA} via the \texttt{LLaMA-Factory} framework\footnote{\url{https://github.com/hiyouga/LLaMA-Factory}}. The primary hyperparameters used for each model are detailed in Table~\ref{tab:hyperparameters}. Common settings across all models include a \texttt{learning\_rate} of $5.0 \times 10^{-5}$, \texttt{bf16} mixed-precision, 4-bit \texttt{bnb} quantization, and a \texttt{cosine} learning rate scheduler.

\begin{table*}[ht]
	\centering
	\caption{Key fine-tuning hyperparameters for all models.}
	\label{tab:hyperparameters}
	
		\begin{tabular}{lccc}
			\toprule
			\textbf{Hyperparameter} & \textbf{Llama-3.1-8B-Instruct} & \textbf{Llama-3.1-70B-Instruct} & \textbf{Phi-4} \\
			\midrule
			\multicolumn{4}{l}{\textit{Model Details}} \\
			\quad \texttt{quantization\_bit} & 4 & 4 & 4 \\
			\quad \texttt{flash\_attn} & \texttt{fa2} & \texttt{fa2} & \texttt{fa2} \\
			\midrule
			\multicolumn{4}{l}{\textit{LoRA Configuration}} \\
			\quad \texttt{lora\_rank} & 32 & 8 & 16 \\
			\quad \texttt{lora\_alpha} & 64 & 32 & 32 \\
			\quad \texttt{lora\_dropout} & 0.05 & 0.05 & 0.05 \\
			\quad \texttt{lora\_target} & \texttt{all} & \texttt{all} & \texttt{all} \\
			\midrule
			\multicolumn{4}{l}{\textit{Training Details}} \\
			\quad \texttt{num\_train\_epochs} & 1.0 & 1.0 & 1.0 \\
			\quad \texttt{warmup\_ratio} & 0.05 & 0.1 & 0.1 \\
			\quad \texttt{per\_device\_train\_batch\_size} & 1 & 1 & 1 \\
			\quad \texttt{gradient\_accumulation\_steps} & 8 & 2 & 8 \\
			\quad \texttt{gradient\_checkpointing} & false & true & false \\
			\bottomrule
		\end{tabular}%
	
\end{table*}


\section{Evaluation Benchmark Details}
\label{app:benchmark_details}

This section provides additional details on the benchmarks used in our experiments (Section~\ref{sec:experiments}).

\paragraph{BFCL-V3:}
The BFCL-V3~\citep{patil2025bfcl} benchmark evaluates multi-turn tool-calling across four challenging categories: \emph{Base}, \emph{Missing Function}, \emph{Missing Parameter}, and \emph{Long Context}. It focuses on conversational scenarios that require clarification, refusal, and robustness. Performance is reported as execution accuracy in a live, executable environment, with credit awarded only when all steps in a dialogue are correctly resolved.

\paragraph{API Bank:}
The API Bank~\citep{li-etal-2023-apibank} benchmark consists of $8$ domains, $73$ tools, and $314$ dialogues. We evaluate on Level 1 (``Call'') and Level 2 (``Retrieve+Call''). As tool retrieval is not the focus of this paper, we provide the set of required tools to the model for both levels. The evaluation is conducted in an off-policy, teacher-forced manner, where the model must predict the next step given the correct history.

\paragraph{CONFETTI:}
The CONFETTI~\citep{alkhouli-etal-2025-confetti} benchmark is a multi-turn, multi-step conversational benchmark with $109$ human-simulated conversations covering $86$ tools. A key characteristic is its explicit inclusion of conversational complexities, such as follow-ups, goal correction, goal switching, ambiguous goals, and over- or under-filled parameters. We focus on evaluating the turn-level tool-calling accuracy.

\newpage
\clearpage
\section{Details of Ablation}
\label{app:ablation_details}
This section provides additional details on the ablation studies discussed in Section~\ref{sec:ablations} of the main paper.

\subsection{Detailed Frontier Model Ablation Results}
\label{app:frontier_results}

Table~\ref{tab:frontier_ablation_full} provides the comprehensive performance breakdown of Llama-3.1-70B fine-tuned on data generated by synthesis models of varying scales and licensing types. This includes detailed metrics for Base, Missing Function, Missing Parameter, and Long Context scenarios.

\begin{table*}[ht]
	\centering
	\caption{Full performance breakdown of Llama-3.1-70B fine-tuned on data generated by different synthesis models.}
	\label{tab:frontier_ablation_full}
		\begin{tabular}{lccccc}
			\toprule
			\textbf{Synthesis Model} & \textbf{Acc} & \textbf{Base} & \textbf{MissFunc} & \textbf{MissParam} & \textbf{LongCtx} \\
			\midrule
			ToolWeave (Mistral-Small-24B) [Open] & 28.00 & 29.00 & 32.50 & 26.50 & 24.00 \\
			ToolWeave (Mistral-Medium) [Commercial] & 28.88 & 27.50 & 39.00 & 30.00 & 19.00 \\
			ToolWeave (GPT-OSS-120B) [Open] & 33.25 & 37.50 & 35.00 & 32.00 & 28.50 \\
			ToolWeave (GPT-5-mini) [Frontier] & \textbf{39.75} & \textbf{42.50} & \textbf{43.00} & \textbf{37.00} & \textbf{36.50} \\
			\bottomrule
		\end{tabular}%
\end{table*}

\subsection{Detailed Component Ablation Results}
\label{app:component_ablations}

This section provides the comprehensive results for the pipeline component ablations discussed in Section~\ref{sec:ablations}.

\paragraph{Necessity of Dialogue Post-Processing:}
Table~\ref{tab:postproc_ablation_full} isolates the impact of the post-processing stage. Injecting robustness patterns yields a substantial \textbf{+6.26\%} absolute gain in Multi-Turn Accuracy. Evaluated on Llama-3.1-8B-Instruct, post-processing produces large and consistent gains across all metrics, particularly in the "Missing Function" category (+7.00).

\begin{table*}[h]
	\centering
	\caption{Performance impact of the post-processing (postproc) stage on Llama-3.1-8B-Instruct.}
	\label{tab:postproc_ablation_full}
	\begin{tabular}{lccccc}
		\toprule
		\textbf{Setting} & \textbf{Acc} & \textbf{Base} & \textbf{MissFunc} & \textbf{MissParam} & \textbf{LongCtx} \\
		\midrule
		ToolWeave (GPT-OSS seed) w/o postproc & 13.62 & 14.50 & 14.50 & 13.50 & 12.00 \\
		ToolWeave (GPT-OSS seed) w/ postproc & \textbf{19.88} & \textbf{23.00} & \textbf{21.50} & \textbf{15.50} & \textbf{19.50} \\
		\midrule
		\textit{Absolute Gain} & \textit{+6.26} & \textit{+8.50} & \textit{+7.00} & \textit{+2.00} & \textit{+7.50} \\
		\bottomrule
	\end{tabular}
\end{table*}

\paragraph{Necessity of Structured Sampling and Fine-Grained Planning:}
Table~\ref{tab:pipeline_ablation_full} validates the architecture of the generation pipeline itself. To ensure comparability, all settings in this table exclude post-processing.
\begin{enumerate}
	\item \textbf{Impact of Structured Sampler:} Replacing our motif-based sampler with a random walk (Row 2) degrades accuracy from 13.62\% to 12.25\%.
	\item \textbf{Impact of Fine-Grained Planner:} Row 3 shows that replacing our planner with ToolFlow's monolithic planner collapses performance to 7.50\%, demonstrating the planner is the most critical component for enabling non-frontier models.
\end{enumerate}

\begin{table*}[h]
	\centering
	\caption{Performance comparison across pipeline settings (excluding post-processing). Validates the Structured Sampler and Fine-Grained Planner.}
	\label{tab:pipeline_ablation_full}
		\begin{tabular}{lccccc}
			\toprule
			\textbf{Setting} & \textbf{Acc} & \textbf{Base} & \textbf{MissFunc} & \textbf{MissParam} & \textbf{LongCtx} \\
			\midrule
			ToolWeave (no postproc) & \textbf{13.62} & \textbf{14.50} & \textbf{14.50} & \textbf{13.50} & \textbf{12.00} \\
			Random sampling + ToolWeave & 12.25 & 13.00 & 14.00 & \textbf{13.50} & 8.50 \\
			Structured sampling + ToolFlow & 7.50 & 9.50 & 7.50 & 5.00 & 8.00 \\
			ToolFlow baseline (GPT-OSS seed) & 7.62 & 11.00 & 4.50 & 5.50 & 9.50 \\
			\bottomrule
		\end{tabular}%
\end{table*}

\subsection{LLM-as-Judge Evaluation Details}
\label{app:llm_judge_details}

This appendix provides full details of the LLM-as-judge protocol used in Section~\ref{sec:llm_dialogue_eval} to assess the semantic quality of synthetic dialogues.

\paragraph{Judge Model and Decoding Setup}
Following \citep{wang2025toolflow}, we use \texttt{Llama-3-405B-Instruct}~\citep{grattafiori2024llama3} as the judge model. To ensure deterministic and reproducible scoring, the model is run with temperature $0.0$, greedy decoding, and a fixed maximum token budget. Each dialogue is evaluated in a single forward pass; no score aggregation or self-consistency sampling is applied.

\paragraph{Evaluation Dimensions}
Each dialogue is independently rated along four dimensions on a 1--5 Likert scale:
\begin{itemize}[leftmargin=*,noitemsep,topsep=0pt,parsep=0pt]
	\item \textbf{Naturalness}: Fluency and human-likeness of the conversation.
	\item \textbf{Coherence}: Logical consistency and relevance across turns.
	\item \textbf{Helpfulness}: Degree to which the dialogue successfully addresses the user’s goal.
	\item \textbf{Accuracy}: Correctness of tool usage, arguments, and factual content.
\end{itemize}

\paragraph{Sampling Protocol}
For each dataset, we randomly sample 200 dialogues, stratified uniformly across the available domains (20 domains for ToolWeave and ToolFlow). ToolDial dialogues are sampled from the publicly released dataset and may not cover all domains. Each sampled dialogue is evaluated exactly once.

\paragraph{Evaluation Prompt}
The judge is provided with a full dialogue transcript and instructed to score it independently along the four dimensions above. Dataset and model identities are not disclosed to the judge. We adopt the full evaluation prompt from \citet{wang2025toolflow}, which is shown below:
\onecolumn
\begin{tcolorbox}[title=LLM-as-judge Evaluation Prompt,
	colback=white, colframe=black!40, fonttitle=\bfseries, breakable, fontupper=\small]
	
    You are asked to evaluate some synthetic dialogue data. These synthetic dialogue occur between the user, the AI assistant, and the tool. Please evaluate the data based on the following criteria, assigning a score from 1 to 5 for each category. Use the detailed descriptions below to guide your assessment:\\

    - Naturalness (1-5 points): Only evaluate whether the user's request and response is natural and realistic. Focus more on the natural flow of the conversation and less on the choice of words. For example, pay attention to whether users will ask similar questions in real scenarios. And assess whether user behavior is natural. For example, real users rarely ask similar questions consecutively or ask longer questions.\\
    
    - Coherence (1-5 points): Evaluate the overall flow and logical connection between the turns in the conversation. Focus on checking whether the user's previous and subsequent rounds of requests are relevant. Also, ensure that meaningful order of agent interactions is maintained. For example, the agent should not provide information before the user requests it. Similarly, the relevant tools must be invoked only after the assistant calls the tool with the relevant parameters. Such discrepancies should be heavily penalized.\\
    
    - Helpfulness (1-5 points): Determine the effectiveness and value of the AI assistant's responses in addressing the user's needs.\\
    
    - Accuracy (1-5 points): Check for the accuracy and consistency of the information provided. Everything returned by the tool is assumed to be accurate. However, at the same time, if the assistant hallucinates or makes up information, it should be heavily penalized.\\
    
    Below are some examples that are rated poorly in each category:\\
    
    1. Naturalness:\\
    - User: "Can you tell me about the weather?"\\
    - Assistant: "The weather is nice."\\
    - User: "What about tomorrow?"\\
    - Assistant: "Tomorrow will be nice too."\\
    
    2. Coherence:\\
    - User: "What's the capital of France?"\\
    - Assistant: "The capital of France is Paris."\\
    - User: "And the capital of Germany?"\\
    - Assistant: "The capital of Germany is Berlin."\\
    
    3. Helpfulness:\\
    - User: "I need help with my homework."\\
    - Assistant: "What subject is it?"\\
    - User: "Math."\\
    - Assistant: "Okay."\\
    
    4. Accuracy:\\
    - User: "What's the capital of Italy?"\\
    - Assistant: "The capital of Italy is Rome."\\
    - User: "What's the capital of Spain?"\\
    - Assistant: "The capital of Spain is Barcelona."\\
    
    Some more examples specifically catered to tool usage:\\
    1. Coherence: (Tool simulated before assistant provides tool to be called along with its input parameters)\\
    - User: "Can you raise a ticket for my internet issue?"\\
    - Tool: \{"ticket\_id": "12345"\}\\
    
    2. Coherence: (Assistant goes on rambling trying to correct erroneous tool output but never calls the tool)\\
    - User: "Please book a flight for me."\\
    - Assistant: "Sure, I can help you with that. Can you please provide me with the details of your flight?"\\
    - User: "I need a flight from New York to San Francisco on March 10th."\\
    - Assistant: book\_flight(\{"from": "New York", "to": "San Francisco", "date": "March 10th"\})\\
    - Tool: "Flight ID not found."\\
    - Assistant: "I apologize for the inconvenience. It seems that there was an error with the flight ID. Let me try again."\\
    - Tool: "Flight ID not found."\\
    
    3. Coherence: (Assistant talks about calling tool but does not use the required tool-calling format)\\
    - User: "Can you check the weather for me?"\\
    - Assistant: "Sure, I can help you with that. Let me check the weather for you."\\
    - Tool: \{"temperature": "72°F", "condition": "Sunny"\}\\
    - Assistant: "The weather is 72°F and sunny."\\
    
    4. Accuracy: (Assistant hallucinates information)\\
    - User: "Can you reserve a table for two at an Italian restaurant in New York City for tomorrow night?"\\
    - Assistant: "Sure, I have reserved a table for two at 'Luigi's Italian Bistro' in New York City for tomorrow night at 7 PM."\\
    
    All such examples must be heavily penalized, and the scores they receive should be 1 or 2.\\
    
    Please use a more **CRITICAL** and **STRICT** evaluation method. After scoring, please provide brief comments or feedback for each category to explain your ratings.\\
    
    Please provide your evaluation in the following format:\\
    
    Evaluation of Synthetic Dialogue Data\\
    
    1. Naturalness: [Score] / 5\\
    - Comments: [Brief comments or feedback]\\
    
    2. Coherence: [Score] / 5\\
    - Comments: [Brief comments or feedback]\\
    
    3. Helpfulness: [Score] / 5\\
    - Comments: [Brief comments or feedback]\\
    
    4. Accuracy: [Score] / 5\\
    - Comments: [Brief comments or feedback]
	
\end{tcolorbox}

\onecolumn
\section{ToolWeave Sample Data - Goal, Plan, Tools, and Dialogue}
\label{app:example}
We present a detailed example from the customer support domain that illustrates a complex, multi-step dialogue. The example showcases the five available tools, the high-level goal constructed using these tools, the 15-step fine-grained plan, and the final multi-turn dialogue that realizes the plan. The following subsections present each component in turn.

\subsection{Tools List}
\label{app:example_toollist}

\begin{tcolorbox}[title=Tool 1: create\_support\_ticket,
	colback=white, colframe=black!20, fonttitle=\bfseries, breakable]	
	\begin{minted}[breaklines=true, fontsize=\small]{json}
		{
			"name": "create_support_ticket",
			"description": "Create support ticket.",
			"parameters": {
				"type": "object",
				"properties": {
					"issue_description": {"type": "string", "description": "A detailed description of the issue."},
					"category": {"type": "string", "enum": ["technical", "billing", "general"], "default": "general"},
					"requester_id": {"type": "string", "description": "The unique identifier for the customer requesting support."},
					"urgency_level": {"type": "string", "enum": ["low", "medium", "high"], "default": "medium"}
				},
				"required": ["issue_description", "requester_id"]
			},
			"results": {
				"type": "object",
				"properties": {
					"ticket_id": {"type": "string", "description": "The unique identifier for the newly created ticket."},
					"creation_date": {"type": "string", "format": "date-time"},
					"status": {"type": "string", "enum": ["open", "in_progress", "resolved", "closed", "on_hold"]}
				}
			}
		}
	\end{minted}
	
\end{tcolorbox}
\newpage

\begin{tcolorbox}[title=Tool 2: get\_ticket\_details,
	colback=white, colframe=black!20, fonttitle=\bfseries, breakable]
	
	\begin{minted}[breaklines=true, fontsize=\small]{json}
		{
			"name": "get_ticket_details",
			"description": "Retrieve comprehensive details about a specific support ticket.",
			"parameters": {
				"type": "object",
				"properties": {
					"support_ticket_identifier": {"type": "string", "description": "The unique reference number of the ticket"}
				},
				"required": ["support_ticket_identifier"]
			},
			"results": {
				"type": "object",
				"properties": {
					"customer_id": {"type": "string"},
					"issue_description": {"type": "string"},
					"priority": {"type": "string", "enum": ["low","medium","high"]},
					"category": {"type": "string", "enum": ["technical","billing","general"]},
					"status": {"type": "string", "enum": ["open","in_progress","resolved","closed","on_hold"]},
					"creation_date": {"type": "string", "format": "date-time"},
					"last_updated": {"type": "string", "format": "date-time"}
				}
			}
		}
	\end{minted}
	
\end{tcolorbox}

\newpage

\begin{tcolorbox}[title=Tool 3: search\_tickets,
	colback=white, colframe=black!20, fonttitle=\bfseries, breakable]
	
	\begin{minted}[breaklines=true, fontsize=\small]{json}
		{
			"name": "search_tickets",
			"description": "Find support tickets by criteria.",
			"parameters": {
				"type": "object",
				"properties": {
					"start_date": {"type": "string", "format": "date"},
					"end_date": {"type": "string", "format": "date"},
					"user_account_id": {"type": "string"},
					"issue_type": {"type": "string", "enum": ["technical","billing","general"], "default": "general"},
					"ticket_state": {"type": "string", "enum": ["open","in_progress","resolved","closed"], "default": "open"},
					"urgency_level": {"type": "string", "enum": ["low","medium","high"], "default": "medium"}
				},
				"required": ["user_account_id"]
			},
			"results": {
				"type": "object",
				"properties": {
					"tickets": {
						"type": "array",
						"items": {
							"type": "object",
							"properties": {
								"ticket_id": {"type": "string"},
								"issue_description": {"type": "string"},
								"creation_date": {"type": "string", "format": "date-time"},
								"last_updated": {"type": "string", "format": "date-time"}
							}
						}
					}
				}
			}
		}
	\end{minted}
	
\end{tcolorbox}
\newpage
\begin{tcolorbox}[title=Tool 4: escalate\_ticket\_to\_specialist,
	colback=white, colframe=black!20, fonttitle=\bfseries, breakable]
	
	\begin{minted}[breaklines=true, fontsize=\small]{json}
		{
			"name": "escalate_ticket_to_specialist",
			"description": "Elevates a support ticket to a specialist team.",
			"parameters": {
				"type": "object",
				"properties": {
					"specialist_team": {"type": "string", "enum": ["technical","billing","legal","management"]},
					"urgency_level": {"type": "string", "enum": ["low","medium","high","critical"], "default": "medium"},
					"attachments": {"type": "array", "items": {"type": "object"}},
					"support_case_id": {"type": "string"},
					"specialist_notes": {"type": "string"}
				},
				"required": ["specialist_team","support_case_id","specialist_notes"]
			},
			"results": {
				"type": "object",
				"properties": {
					"escalation_id": {"type": "string"},
					"escalation_date": {"type": "string", "format": "date-time"},
					"status": {"type": "string", "enum": ["pending","in_progress","resolved","rejected"]}
				}
			}
		}
	\end{minted}
	
\end{tcolorbox}

\begin{tcolorbox}[title=Tool 5: update\_escalation\_status,
	colback=white, colframe=black!20, fonttitle=\bfseries, breakable]
	
	\begin{minted}[breaklines=true, fontsize=\small]{json}
		{
			"name": "update_escalation_status",
			"description": "Modify ticket escalation status.",
			"parameters": {
				"type": "object",
				"properties": {
					"status": {"type": "string", "enum": ["pending","in_progress","resolved","rejected"]},
					"ticket_escalation_id": {"type": "string"}
				},
				"required": ["status","ticket_escalation_id"]
			},
			"results": {
				"type": "object",
				"properties": {
					"last_updated": {"type": "string", "format": "date-time"}
				}
			}
		}
	\end{minted}
	
\end{tcolorbox}

\subsection{High Level Goal}
\label{app:example_goal}
\begin{tcolorbox}[title=Goal, colback=white, colframe=black!40, fonttitle=\bfseries, breakable]
	\textit{Create a support ticket, review its details, and escalate it to a specialist team for urgent resolution.}
\end{tcolorbox}

\subsection{Plan}
\label{app:example_plan}
\begin{tcolorbox}[title=Multi-step Plan, colback=white, colframe=black!40, fonttitle=\bfseries, breakable]
	
	\noindent\textcolor{usercolor}{\textbf{Step 1 (USER\_UTTERANCE)}}\\
	Utterance: \textit{I need to create a support ticket and get all the details about it.}\\
	Provided params:
	\begin{minted}[breaklines=true, fontsize=\small]{json}
		{
			"create_support_ticket.requester_id":
			"$user_provided_$create_support_ticket.requester_id"
		}
	\end{minted}
	
	\medskip
	\noindent\textcolor{assistantcolor}{\textbf{Step 2 (ASSISTANT\_CLARIFICATION)}}\\
	Parameter names:
	\begin{minted}[breaklines=true, fontsize=\small]{json}
		["create_support_ticket.issue_description"]
	\end{minted}
	
	\medskip
	\noindent\textcolor{usercolor}{\textbf{Step 3 (USER\_RESPONSE\_TO\_CLARIFICATION)}}\\
	Provides params:
	\begin{minted}[breaklines=true, fontsize=\small]{json}
		{
			"create_support_ticket.issue_description":
			"$user_provided_$create_support_ticket.issue_description"
		}
	\end{minted}
	
	\medskip
	\noindent\textcolor{callcolor}{\textbf{Step 4 (CALL\_TOOL: create\_support\_ticket)}}\\
	Parameters:
	\begin{minted}[breaklines=true, fontsize=\small]{json}
		{
			"create_support_ticket.issue_description":
			"$user_provided_$create_support_ticket.issue_description",
			"create_support_ticket.requester_id":
			"$user_provided_$create_support_ticket.requester_id"
		}
	\end{minted}
	
	\medskip
	\noindent\textcolor{callcolor}{\textbf{Step 5 (CALL\_TOOL: get\_ticket\_details)}}\\
	Parameters:
	\begin{minted}[breaklines=true, fontsize=\small]{json}
		{
			"get_ticket_details.support_ticket_identifier":
			"$create_support_ticket.ticket_id"
		}
	\end{minted}
	
	\medskip
	\noindent\textcolor{assistantcolor}{\textbf{Step 6 (ASSISTANT\_RESPONSE\_TOOL)}}\\
	Summarizes tools:
	\begin{minted}[breaklines=true, fontsize=\small]{json}
		["create_support_ticket", "get_ticket_details"]
	\end{minted}
	Outputs:
	\begin{minted}[breaklines=true, fontsize=\small]{json}
        [
            "create_support_ticket.ticket_id",
            "create_support_ticket.creation_date",
            "create_support_ticket.status",
            "get_ticket_details.customer_id",
            "get_ticket_details.issue_description",
            "get_ticket_details.priority",
            "get_ticket_details.category",
            "get_ticket_details.status",
            "get_ticket_details.creation_date",
            "get_ticket_details.last_updated"
        ]
	\end{minted}
	
	\medskip
	\noindent\textcolor{usercolor}{\textbf{Step 7 (USER\_UTTERANCE)}}\\
	Utterance: \textit{Can you search for the ticket I just created?}
	
	\medskip
	\noindent\textcolor{callcolor}{\textbf{Step 8 (CALL\_TOOL: search\_tickets)}}\\
	Parameters:
	\begin{minted}[breaklines=true, fontsize=\small]{json}
		{
			"search_tickets.user_account_id": "$get_ticket_details.customer_id",
			"search_tickets.issue_type": "$get_ticket_details.category",
			"search_tickets.ticket_state": "$create_support_ticket.status"
		}
	\end{minted}
	
	\medskip
	\noindent\textcolor{assistantcolor}{\textbf{Step 9 (ASSISTANT\_RESPONSE\_TOOL)}}\\
	Summarizes tools:
	\begin{minted}[breaklines=true, fontsize=\small]{json}
		["search_tickets"]
	\end{minted}
	Outputs:
	\begin{minted}[breaklines=true, fontsize=\small]{json}
        [
            "search_tickets.tickets[].ticket_id",
            "search_tickets.tickets[].issue_description",
            "search_tickets.tickets[].creation_date",
            "search_tickets.tickets[].last_updated"
        ]
	\end{minted}
	
	\medskip
	\noindent\textcolor{usercolor}{\textbf{Step 10 (USER\_UTTERANCE)}}\\
	Utterance: \textit{Now, please escalate this ticket to a specialist team for urgent resolution and update the escalation status.}\\
	Provided params:
	\begin{minted}[breaklines=true, fontsize=\small]{json}
		{
			"escalate_ticket_to_specialist.specialist_team":
			"$user_provided_$escalate_ticket_to_specialist.specialist_team"
		}
	\end{minted}
	
	\medskip
	\noindent\textcolor{assistantcolor}{\textbf{Step 11 (ASSISTANT\_CLARIFICATION)}}\\
	Parameter names:
	\begin{minted}[breaklines=true, fontsize=\small]{json}
        [
            "escalate_ticket_to_specialist.specialist_notes",
            "update_escalation_status.status"
        ]
	\end{minted}
	
	\medskip
	\noindent\textcolor{usercolor}{\textbf{Step 12 (USER\_RESPONSE\_TO\_CLARIFICATION)}}\\
	Provides params:
	\begin{minted}[breaklines=true, fontsize=\small]{json}
        {
            "escalate_ticket_to_specialist.specialist_notes":
            "$user_provided_$escalate_ticket_to_specialist.specialist_notes",
            "update_escalation_status.status":
            "$user_provided_$update_escalation_status.status"
        }
	\end{minted}
	
	\medskip
	\noindent\textcolor{callcolor}{\textbf{Step 13 (CALL\_TOOL: escalate\_ticket\_to\_specialist)}}\\
	Parameters:
	\begin{minted}[breaklines=true, fontsize=\small]{json}
        {
            "escalate_ticket_to_specialist.specialist_team":
            "$user_provided_$escalate_ticket_to_specialist.specialist_team",
            "escalate_ticket_to_specialist.support_case_id":
            "$create_support_ticket.ticket_id",
            "escalate_ticket_to_specialist.specialist_notes":
            "$user_provided_$escalate_ticket_to_specialist.specialist_notes"
        }
	\end{minted}
	
	\medskip
	\noindent\textcolor{callcolor}{\textbf{Step 14 (CALL\_TOOL: update\_escalation\_status)}}\\
	Parameters:
	\begin{minted}[breaklines=true, fontsize=\small]{json}
        {
            "update_escalation_status.status":
            "$user_provided_$update_escalation_status.status",
            "update_escalation_status.ticket_escalation_id":
            "$escalate_ticket_to_specialist.escalation_id"
        }
	\end{minted}
	
	\medskip
	\noindent\textcolor{assistantcolor}{\textbf{Step 15 (ASSISTANT\_RESPONSE\_TOOL)}}\\
	Summarizes tools:
	\begin{minted}[breaklines=true, fontsize=\small]{json}
		["escalate_ticket_to_specialist", "update_escalation_status"]
	\end{minted}
	Outputs:
	\begin{minted}[breaklines=true, fontsize=\small]{json}
        [
            "escalate_ticket_to_specialist.escalation_id",
            "escalate_ticket_to_specialist.escalation_date",
            "escalate_ticket_to_specialist.status",
            "update_escalation_status.last_updated"
        ]
	\end{minted}
	
\end{tcolorbox}

\subsection{Dialogue}
\label{app:example_dialog}
\begin{tcolorbox}[title=Multi-turn Dialogue, colback=white, colframe=black!40, fonttitle=\bfseries, breakable]
	
	\systemline{Current time: 2025-08-27 21:24:05.}
	
	\medskip
	\userline{I need to create a support ticket for an issue I'm having with my account login. I keep getting an error message saying 'Invalid credentials' even though I'm sure I'm entering the correct username and password. I've tried resetting my password multiple times, but the issue persists. I need this resolved as soon as possible because I can't access any of my account features.}
	
	\medskip
	\assistantline{Could you please provide your requester ID so I can create a support ticket for your account login issue?}
	
	\medskip
	\userline{My requester ID is cust123456789.}
	
	\medskip
	\toolcall{create\_support\_ticket(\{"issue\_description": "I keep getting an error message saying 'Invalid credentials' even though I'm sure I'm entering the correct username and password. I've tried resetting my password multiple times, but the issue persists. I need this resolved as soon as possible because I can't access any of my account features.", "requester\_id": "cust123456789", "urgency\_level": "high", "category": "general"\})}
	
	\medskip
	\toolresp{\{"ticket\_id": "tkt987654321", "creation\_date": "2025-08-27T21:24:05Z", "status": "open"\}}
	
	\medskip
	\toolcall{get\_ticket\_details(\{"support\_ticket\_identifier": "tkt987654321"\})}
	
	\medskip
	\toolresp{\{"customer\_id": "cust123456789", "issue\_description": "I keep getting an error message saying 'Invalid credentials' even though I'm sure I'm entering the correct username and password. I've tried resetting my password multiple times, but the issue persists. I need this resolved as soon as possible because I can't access any of my account features.", "priority": "high", "category": "general", "status": "open", "creation\_date": "2025-08-27T21:24:05Z", "last\_updated": "2025-08-27T21:24:05Z"\}}
	
	\medskip
	\assistantline{I've created a support ticket for your account login issue. Your ticket ID is tkt987654321, created on August 27, 2025, and it is currently open. The ticket details confirm that the issue is related to receiving an 'Invalid credentials' error message despite entering the correct username and password, and that you've attempted to reset your password multiple times without success. The ticket is categorized as 'general' with high priority. The ticket status remains open, and it was last updated on August 27, 2025. Our support team will address this as soon as possible.}
	
	\medskip
	\userline{Can you search for the ticket I just created?}
	
	\medskip
	\toolcall{search\_tickets(\{"user\_account\_id": "cust123456789", "issue\_type": "general", "ticket\_state": "open", "urgency\_level": "medium"\})}
	
	\medskip
	\toolresp{\{"tickets": [\{"ticket\_id": "tkt987654321", "issue\_description": "I keep getting an error message saying 'Invalid credentials' even though I'm sure I'm entering the correct username and password. I've tried resetting my password multiple times, but the issue persists. I need this resolved as soon as possible because I can't access any of my account features.", "creation\_date": "2025-08-27T21:24:05Z", "last\_updated": "2025-08-27T21:24:05Z"\}]\}}
	
	\medskip
	\assistantline{I found your open support ticket with ID tkt987654321. It was created on August 27, 2025, regarding the 'Invalid credentials' error you're experiencing despite entering the correct username and password. The issue description mentions that you've tried resetting your password multiple times without success. This ticket was last updated on August 27, 2025, and our support team is working to resolve it as quickly as possible.}
	
	\medskip
	\userline{Please escalate my ticket to a specialist team for urgent resolution. Here are some additional details: I've been experiencing this issue for the past three days, and it's preventing me from accessing critical account features. I've tried all the standard troubleshooting steps, including clearing my browser cache and trying different devices, but nothing has worked. I suspect there might be an issue with the account authentication system itself.}
	
	\medskip
	\assistantline{To which specialist team would you like me to escalate your ticket for urgent resolution?}
	
	\medskip
	\userline{Please escalate my ticket to the technical team and set the escalation status to in\_progress.}
	
	\medskip
	\toolcall{escalate\_ticket\_to\_specialist(\{"specialist\_team": "technical", "support\_case\_id": "tkt987654321", "specialist\_notes": "I've been experiencing this issue for the past three days, and it's preventing me from accessing critical account features. I've tried all the standard troubleshooting steps, including clearing my browser cache and trying different devices, but nothing has worked. I suspect there might be an issue with the account authentication system itself.", "attachments": [], "urgency\_level": "high"\})}
	
	\medskip
	\toolresp{\{"escalation\_id": "esc987654321", "escalation\_date": "2025-08-27T21:24:05Z", "status": "in\_progress"\}}
	
	\medskip
	\toolcall{update\_escalation\_status(\{"status": "in\_progress", "ticket\_escalation\_id": "esc987654321"\})}
	
	\medskip
	\toolresp{\{"last\_updated": "2025-08-27T21:24:05Z"\}}
	
	\medskip
	\assistantline{Your ticket has been successfully escalated to the technical team. The escalation ID is esc987654321, created on August 27, 2025, with a status of "in\_progress". This escalation was last updated on August 27, 2025. The technical team will now prioritize resolving your account login issue, considering the additional details you've provided about the problem's duration and your troubleshooting attempts.}
	
\end{tcolorbox}


\section{Example Dialogue Generated via ToolFlow}
\label{app:toolflow_example}
This example dialogue highlights a common issue in dialogues generated via the ToolFlow pipeline, where the tool agent hallucinates a tool response without the assistant agent providing an actual tool call, resulting in a fabricated tool reply (as seen in the first assistant response and the first tool response).

\begin{tcolorbox}[title=Multi-turn Dialogue, colback=white, colframe=black!40, fonttitle=\bfseries, breakable]
	
	\userline{I've just bought a new field and have the map outline. 
Could you define its boundary with this GeoJSON: 
\texttt{\{"type":"Feature",\allowbreak "geometry":\{"type":"Polygon",\allowbreak
"coordinates":[[[30.0,10.0],\allowbreak [40.0,40.0],\allowbreak
[20.0,40.0],\allowbreak [10.0,20.0],\allowbreak [30.0,10.0]]]\},\allowbreak
"properties":\{\}\}}? 
The field ID is "field-12345".}

	\medskip
	\assistantline{Sure, defining the boundary for field field-12345 now.}

    \medskip
    \toolresp{\{"field\_boundary\_id": "boundary-67890", "status": "defined"\}}

    \medskip
	\assistantline{Great! The boundary for field field-12345 has been successfully defined (status: defined). I can now pull a short-term weather forecast for the next week for this field. Would you like me to retrieve that for you?}

    \medskip
    \userline{Yes, please pull the short-term weather forecast for the next week for field-12345.}

    \medskip
    \assistantline{Sure! To get the short-term weather forecast I'll need the farm identifier that corresponds to this field. Could you provide the farm ID (or let me know if the field ID can be used in place of a farm ID)?}

    \medskip
    \userline{Sure, the farm ID is farm-abc123.}

    \medskip
    \toolcall{get\_weather\_forecast(\{"farm\_identifier": "farm-abc123"\})}

    \medskip
    \toolresp{\{"temperature\_celsius": 22.3, "precipitation\_mm": 0.0, "wind\_speed\_kph": 14.5, "condition": "sunny"\}}

    \medskip
    \assistantline{Here's the forecast I've retrieved for tomorrow at your field field-12345 (farm-abc123):}
	
\end{tcolorbox}

\newpage

\section{Some Example Dialogues From ToolDial Training Set}
\label{app:tooldial_example_dialogues}

The following examples illustrate some of the issues why LLMs fine-tuned on ToolDial data exhibit inferior performance on the BFCL-V3 benchmark (see Table~\ref{tab:multi_turn_results}). 

Example~\ref{app:tooldial_sample_dialogue_1} shows that models trained on such data struggle with parameter elicitation and error recovery, the assistant neither asks for the missing hotel IP address nor attempts to resolve the misunderstanding after the user's correction.

\subsection{Example 1}
\label{app:tooldial_sample_dialogue_1}
\begin{tcolorbox}[title=Multi-turn Dialogue, colback=white, colframe=black!40, fonttitle=\bfseries, breakable]
	
	\userline{Can you help me find detailed information about local businesses, such as restaurants and cafes, near my hotel? I only have the hotel's IP address and need to find its geo-location first to get the coordinates.}
	
	\medskip
	\assistantline{I can help you find detailed information about local businesses near your hotel. Would you like me to proceed with finding nearby businesses?}

    \medskip
    \userline{No, that's not what I need.}

    \medskip
    \assistantline{I'm sorry, but I cannot fulfill your request due to insufficient information. Please provide more details or try a different query.}
	
\end{tcolorbox}


\subsection{Example 2}
\label{app:tooldial_sample_dialogue_2}
Example~\ref{app:tooldial_sample_dialogue_2} illustrates two issues. The user's initial request is somewhat unrealistic and unlikely to occur in a real-world setting. The assistant's clarification questions are phrased using internal parameter names rather than natural, intuitive language that a user would expect. 

\begin{tcolorbox}[title=Multi-turn Dialogue, colback=white, colframe=black!40, fonttitle=\bfseries, breakable]

	\medskip
	\userline{What is the population within a defined area around a city, given only the city's name and its country?}

	\medskip
	\assistantline{Would it be appropriate to retrieve the population data within a defined area around the city you mentioned?}

	\medskip
	\userline{That would be great. Sure, that sounds good.}

	\medskip
	\assistantline{Please tell me the radius of the circular area around the coordinates and the coordinates of the geographic point to search nearby.}

	\medskip
	\userline{I can provide the radius, which is 300, but I don't know the coordinates.}

	\medskip
	\assistantline{Please provide the namePrefixDefaultLangResults for the city.}

	\medskip
	\userline{Sure! The namePrefixDefaultLangResults for the city is San Francisco.}

	\medskip
	\toolcall{[\{\ "name": "cities", "arguments": \{\ "namePrefixDefaultLangResults": "San Francisco" \}\ \}\ ]}

	\medskip
	\toolresp{\{\ "error": "", "response": \{\ "data": [\{\ "id": 5392171, "wikiDataId": "Q123456", "type": "CITY", "city": "San Francisco", "name": "San Francisco", "country": "United States", "countryCode": "US", "region": "California", "regionCode": "CA", "regionWdId": "Q99", "population": 883305, "coordinates": "[37.7749, -122.4194]"\}, \{\ "id": 5902958, "wikiDataId": "Q123457", "type": "CITY", "city": "San Jose", "name": "San Jose", "country": "United States", "countryCode": "US", "region": "California", "regionCode": "CA", "regionWdId": "Q99", "population": 1035317, "coordinates": "[37.3382, -121.8863]"\}, \{\ "id": 5322760, "wikiDataId": "Q123458", "type": "CITY", "city": "San Diego", "name": "San Diego", "country": "United States", "countryCode": "US", "region": "California", "regionCode": "CA", "regionWdId": "Q99", "population": 1423851, "coordinates": "[32.7157, -117.1611]"\}], "links": \{\ "next": "/v1/geo/cities?offset=60\&limit=50" \}, "metadata": \{\ "currentOffset": 10, "totalCount": 150 \}\ \}\ \}}

	\medskip
	\assistantline{Please choose one of the following cities: San Francisco, San Jose, San Diego.}

	\medskip
	\userline{I choose San Francisco.}

	\medskip
	\toolcall{get\_population(\{\ "radius": "300", "coordinate": "[37.7749, -122.4194]" \})}

	\medskip
	\toolresp{\{\ "error": "", "response": \{\ "radius": 300, "population": 12000, "description": "The estimated resident population within a 300-meter radius of the specified coordinates in San Francisco is approximately 12,000 individuals. This figure is based on the latest demographic data and reflects the density of the area, which includes residential buildings, parks, and commercial establishments.", "coordinates": [37.7749, -122.4194] \}\ \}}

	\medskip
	\assistantline{The estimated resident population within a 300-meter radius of the specified coordinates in San Francisco is approximately 12,000 individuals. This figure is based on the latest demographic data and reflects the density of the area, which includes residential buildings, parks, and commercial establishments.}

	\medskip
	\userline{Thank you for the information. Goodbye!}

	\medskip
	\assistantline{You're welcome! Goodbye!}

\end{tcolorbox}